\documentclass[lettersize, journal]{IEEEtran}

\usepackage{amsmath,amsfonts}
\usepackage{algpseudocode}
\usepackage{algorithm}
\usepackage{array}
\usepackage{textcomp}
\usepackage{stfloats}
\usepackage{url}
\usepackage{verbatim}
\usepackage{graphicx}
\usepackage{cite}

\usepackage{amssymb}
\usepackage{mathrsfs}
\usepackage{graphicx}
\usepackage{xcolor}
\usepackage{booktabs} 
\usepackage{tabularx}
\usepackage{float}
\usepackage{multirow}
\usepackage{threeparttable}
\usepackage{subfigure}
\usepackage{subcaption}
\usepackage{url}

\def\BibTeX{{\rm B\kern-.05em{\sc i\kern-.025em b}\kern-.08em
    T\kern-.1667em\lower.7ex\hbox{E}\kern-.125emX}}
\begin{document}

\title{Can You Do Both? Balancing Order Serving and Crowdsensing for Ride-Hailing Vehicles \textcolor{red}{TBD}}
\title{Online Location Planning for AI-Defined Vehicles: Optimizing Joint Tasks of Order Serving and Spatio-Temporal Heterogeneous Model Fine-Tuning}

\author{
	Bokeng~Zheng*, Bo~Rao*, 
        Tianxiang~Zhu,
        Chee Wei Tan,~\IEEEmembership{Member,~IEEE,}
        Jingpu~Duan,~\IEEEmembership{Member,~IEEE,}
        Zhi~Zhou,~\IEEEmembership{Member,~IEEE,}
        Xu~Chen,~\IEEEmembership{Senior Member,~IEEE,}
        Xiaoxi~Zhang,~\IEEEmembership{Member,~IEEE}

 
        
\thanks{Boken Zheng, B. Rao, T. Zhu, Z. Zhou, X. Chen, and X. Zhang are with the School of Computer Science and Engineering, Sun Yat-sen University, Guangzhou 510006, China~(E-mail: \{zhengbk6,raob5,zhutx23\}@mail2.sysu.edu.cn; \{zhouzhi9,chenxu35,zhangxx89\}@mail.sysu.edu.cn).}
\thanks{Chee Wei Tan is with the College of Computing and Data Science, Nanyang Technological University, Singapore (E-mail: cheewei.tan@ntu.edu.sg).}
\thanks{J. Duan is with the Department of Communications, Peng Cheng Laboratory, Shenzhen 518066, China (E-mail: duanjp@pcl.ac.cn).}
\thanks{* indicates the co-first authors; Xiaoxi Zhang is the corresponding author.}
\thanks{A previous version appears at IWQoS 2024 as a short paper. 
This journal article contains substantial new results about the scenario, problem formulation, and experiments.
}
}


\IEEEpubidadjcol
    \IEEEoverridecommandlockouts
\maketitle

\begin{abstract}
Advances in artificial intelligence (AI) including foundation models (FMs), are increasingly transforming human society, with smart city driving the evolution of urban living. 
Meanwhile, vehicle crowdsensing (VCS) has emerged as a key enabler, leveraging vehicles' mobility and sensor-equipped capabilities. In particular, ride-hailing vehicles can effectively facilitate flexible data collection and contribute towards urban intelligence, despite resource limitations. Therefore, this work explores a promising scenario, where edge-assisted vehicles perform joint tasks of order serving and the emerging foundation model fine-tuning using various urban data.
However, integrating the VCS AI task with the conventional order serving task is challenging, due to their inconsistent spatio-temporal characteristics: (i) The distributions of ride orders and data point-of-interests (PoIs) may not coincide in geography, both following a priori unknown patterns; (ii) they have distinct forms of temporal effects, i.e., prolonged waiting makes orders become instantly invalid while data with increased staleness gradually reduces its utility for model fine-tuning.
To overcome these obstacles, we propose an online framework based on multi-agent reinforcement learning (MARL) with careful augmentation. A new quality-of-service (QoS) metric is designed to characterize and balance the utility of the two joint tasks, under the effects of varying data volumes and staleness. We also integrate graph neural networks (GNNs) with MARL to enhance state representations, capturing graph-structured, time-varying dependencies among vehicles and across locations. Extensive experiments on our testbed simulator, utilizing various real-world foundation model fine-tuning tasks and the New York City Taxi ride order dataset, demonstrate the advantage of our proposed method.
\end{abstract}




\section{Introduction}
\begin{figure}[ht]
\centerline{\includegraphics[width=1.0\linewidth]{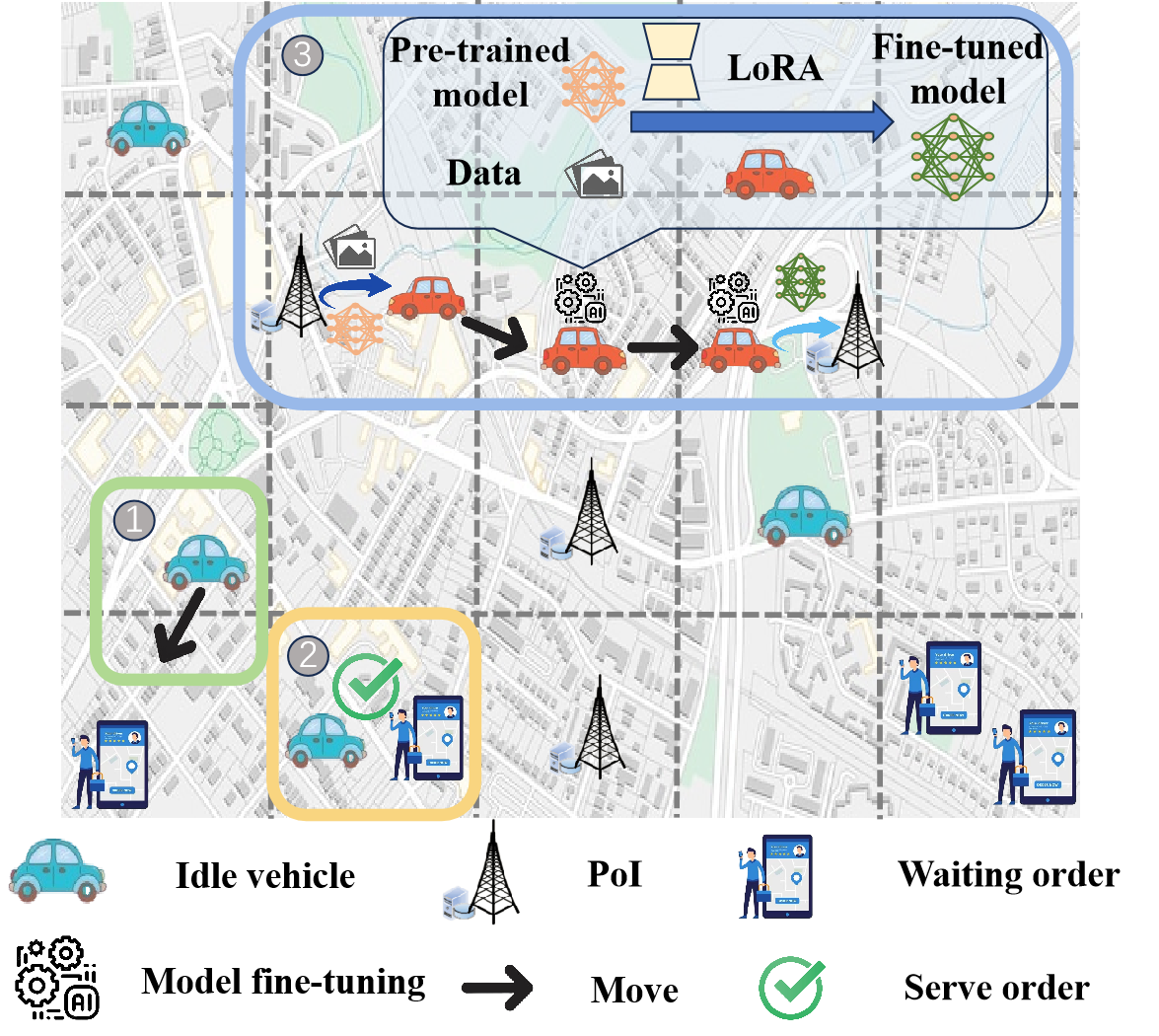}}
\caption{
An illustration of the proposed vehicle scheduling framework that integrates order fulfillment and model fine-tuning tasks in a smart city environment: Box 1 shows idle vehicles dispatched to neighboring areas in preparation for future tasks, while Box 2 illustrates vehicles providing services to order customers. Box 3 depicts the process where a vehicle retrieves the latest model and data from the nearest RSU, fine-tunes the model during movement, and uploads the updated model to a nearby RSU. Each RSU, equipped with a server, stores a complete base model, enabling vehicles to perform real-time fine-tuning as they collect data and transfer the refined models to other RSUs.
}
\label{fig:system overview}
\end{figure}



With the rapid growth of urban population worldwide, urban management faces increasing challenges, giving rise to the concept of smart cities, which aim at improving urban lives through environmental monitoring \cite{EnvironmentalPollution}, traffic control \cite{PervasiveandMobileComputing}, healthcare \cite{ExpertSystems}, etc. Urban data is essential for smart city applications, and how to obtain and use data effectively is a key issue. Fortunately, mobile crowdsensing (MCS) \cite{ITJ19} provides a useful way to collect data, making use of users' mobile devices (or users themselves) as sensing units to complete large-scale and complex social sensing tasks. Compared to MCS, the drawbacks of using fixed devices for data collection include mainly high installation and maintenance costs, as well as very limited coverage \cite{BigData}. 
%
Building on the success of MCS, vehicle crowd-sensing (VCS) present unprecedented opportunitiesleveraging the mobility of vehicles, including unmanned and electric types, equipped with high-precision sensors, can collect various types of data, such as air quality data, traffic conditions, and street view images, to assist government agencies in better city management. 
Notably, ride-hailing vehicles are particularly advantageous for VCS tasks, due to their centralized ride-hailing platform management, which reduces the cost of deploying and executing crowd-sensing tasks, and utilizes the data and computing resources from ride-hailing vehicles to maximize the VCS task utilities.

\noindent {\bf Opportunities and challenges.} In the meanwhile, foundation model (FM)-powered AI applications have revolutionized numerous aspects of human lives, including healthcare, education, industry, etc. FMs, e.g., BERT, GPT-4, ViT, serve as foundation for different downstream tasks in languages, vision, graph processing, and multimodal applications. Among them, a notable emerging category is urban foundation models (UFMs), which are trained on diverse modalities of urban data and are designed to interpret them effectively. UFMs supported applications include environmental monitoring, urban planning, energy management, and our envisioned future in-vehicle infotainment (IVI) applications\cite{you2024raccoon}. 
While a substantial body work has developed various UFMs and downstream tasks~\cite{zhang2024urban},~\cite{mai2023opportunities},~\cite{lu2024ai},~\cite{gao2024survey},
optimization-based strategies that leverage existing infrastructures of MCS and ride-hailing vehicles to {\em execute} UFM-based tasks for {\em utility maximization}, along with fulfilling traditional order serving tasks, are under-explored. This is mainly due to: (i) UFMs trained by various, multi-modal urban data still require fine-grained adaptation for distinct downstream urban tasks; but fine-tuning using spatio-temporal heterogeneous data and performed by a large number of moving vehicles typically needs complex quantitative modeling to maximize the VCS utility. (ii) Deploying model fine-tuning has to be balanced with the conventional order tasks of ride-hailing vehicles. 

\begin{figure}[ht]
\centering
\begin{minipage}[t]{0.48\linewidth}
  \centering
  \includegraphics[width=\linewidth]{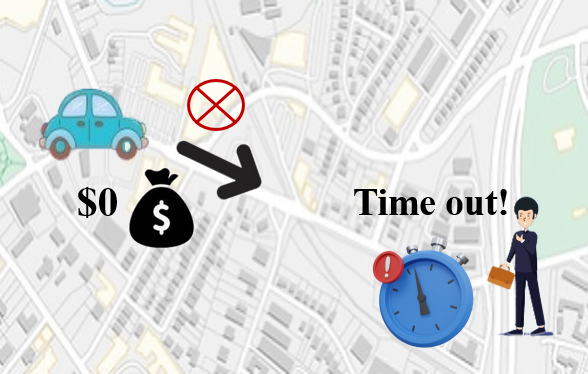}
  \caption*{(a)}  
\end{minipage}%
\hfill
\begin{minipage}[t]{0.48\linewidth}
  \centering
  \includegraphics[width=\linewidth]{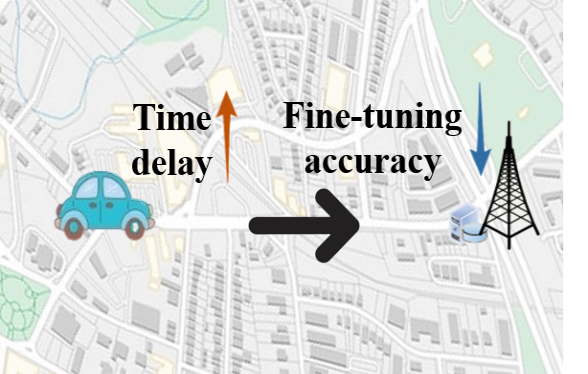}
  \caption*{(b)}  
\end{minipage}
\caption{
(a) Illustrates how an order request that is not completed within the specified deadline results in zero utility for the vehicle.
(b) Depicts how an increasing delay in data collection progressively diminishes the inference accuracy when such data is used for model fine-tuning.}
\label{fig:impact_on_model}
\end{figure}

\noindent {\bf Model Fine-Tuning and Order Serving Tasks Scenario.} In this work, we take the first attempt to comprehensively study the scenario where ride-hailing vehicles participate in VCS tasks using parameter efficient fine tuning (PEFT) techniques~\cite{han2024parameter} to adapt distinct UFMs, while maintaining the opportunities to pick up passengers for order serving. As illustrated in Fig.~\ref{fig:system overview}, each vehicle can cruise in the city and gain monetary utilities in both tasks. There are a set of data points of interest (PoI) distributed in cities, which represent the locations of RSUs equipped with edge servers, where any one or more types of labeled dataset can be generated in real-time nearby and stored. Due to the large volume, data stored in the RSU server can be discarded in a certain period of time. In practice, these data can be descriptive features and feedbacks (labels) of recommendation or generative AR applications, generated by nearby visitors or residents. They can also be traffic/environment monitoring data with labels generated by running efficient deep learning model inference deployed in the RSU. The government or any company that collaborates with the ride-hailing vehicle company has multiple types of VSC tasks to fulfill, each of which needs certain locations of data for fine-tuning UFMs. Whenever a vehicle chooses to perform a VSC task, it needs to download the corresponding UFM from the cloud, move to associated locations for collect real-time data, fine-tune the Low-Rank Adapters (LoRA) of the UFM on the route, and push the fine-tuned adapters to a nearby RSU or the cloud for the task owner to use in UFM updates and inference.
Meanwhile, passengers' order requests also emerge at various times and locations, waiting for ride-hailing vehicles to serve. Therefore, at any given time, each vehicle can cruise to a certain location (squared grid), pick up the passengers to serve the orders before order acceptance deadlines, and/or collect data from a PoI along the route, or just stay in its current place. As depicted in Fig.~\ref{fig:impact_on_model}, both delays in order acceptance and in data collection for fine-tuning can significantly impact the vehicle's utility. Our goal is to provide best next-step location decisions on behalf of each vehicle, to maximize the overall utility of all the vehicles in performing the joint tasks of order serving and UFM model fine-tuning. 

Despite promising benefit that can be brought to smart city development by leveraging UFMs, PETF techniques, and the advantages (mobility, resources) of ride-hailing vehicles, there are three types of technical challenges that we need to address. 

\begin{itemize}
\item{{\bf High-dimensional states in vehicle location planning for spatio-temporal heterogeneous UFM fine-tuning.} 
In this work, our goal is to maximize the long-term, overall utility of all the tasks that fulfilled by the ride-hailing vehicles. Given the uncertain effects of different factors on the performance of model fine-tuning using urban data, a natural and advantageous choice of optimization framework is reinforcement learning (RL), which can effectively learn the optimal online decisions through sequential interactions with dynamic environments.
However, employing global information for centralized decision-making frequently results in state space expansion and poor scalability. Distributed decision-making methods, e.g, multi-agent RL (MARL), prove more apt in addressing these issues when compared to centralized approaches. But an associated challenge arises, i.e., the optimization performance can be suboptimal, resulting from independent decisions based solely on local states. Moreover, the widely adopted performance metric for UFM-based downstream is inference accuracy, but the controllability of our framework is in vehicles' locations and their choices of tasks. The underline mapping between our decisions and the final model accuracy can be affected by various factors in uncertain forms. This results in complex, high-dimensional states, and thus potentially slow convergence of MARL algorithms.} 

\smallskip
\item{{\bf Inconsistent geo-distributions of data PoIs and orders.} 
Ride-hailing orders are mostly concentrated in certain areas of the city, such as the central business district. There may be very few ride-hailing orders in remote areas, such as the outskirts of the city. Unlike order-serving, many UFM based VCS tasks require vehicles to collect data from every corner of the city~\cite{ding2021multi}, ~\cite{chen2016intelligent}, ~\cite{luo2019data}.
 There is a potential scenario where increased attention is directed towards PoIs located in remote suburban areas. An example of such a scenario is the collection of air quality data from environmental monitoring stations situated in forested regions. Collecting data distributed in areas beyond the city center in a timely manner becomes challenging when vehicles are overly focused on seeking ride orders and concentrated in the city center. To simultaneously enhance overall the efficiency of order fulfillment and data collection, it is imperative to strategically allocate vehicles among different regions based on the distribution of orders and data points. In order to increase the effectiveness of MARL algorithms, it is crucial to select the right association between an algorithmic ``agent'' and an environmental item (a location, a vehicle, or a task). Besides, adopting MARL may fall short of handling this inconsistent geo-distributions and balancing the vehicles onto different location, as each agent's decisions may only focus on its local perspective, while allowing full environmental states for each vehicle hinders the scalability and violates our original goal to mitigate the high-dimensionality of state space.}

\smallskip
\item{{\bf Different utility characteristics and requirements of order-serving and model fine-tuning.} 
Since each order-serving task has a pickup deadline and a fixed route determined exogenously (e.g., shortest path or minimum travel time), the objective is to maximize high-value order completions within certain distance constraints. In contrast, fine-tuning FMs requires vehicles to collect fine-tuning data of sufficient {\em volume and quality}. Achieving sufficient data volume may involve guiding the RL agent to visit data-rich locations or multiple locations for broader collection. Data quality factors may further include similarity to the base foundation model's training data and alignment with downstream tasks' inference data. 
Moreover, order serving and data collection tasks have different {\em time-sensitivity}. Long delays in response to an order prompt the passenger to cancel the request, rendering it an invalid order. Differently, since fine-tuning data is generated in real time, it possesses a characteristic of {\em data freshness} which should effect the model performance in a fine-grained manner. For instance, inference accuracy of a fine-tuned model probably benefits from vehicles that promptly reach data collection points and quickly update fine-tuned PEFT adapters to minimize latency between fine-tuning and inference. These varying demands add complexity to RL agents' learning processes in balancing the gains of the two jointly executed tasks.}

\end{itemize}
To address the above identified technical challenges simultaneously, we propose an online optimization framework based on multi-agent reinforcement learning, fusing several customized techniques, in particular, to handle the complex structure introduced by vehicles' mobility, spatio-temporal heterogeneity in environmental data, as well as the coupled and possibly conflicting interests in performing the two types of tasks. 
To achieve this, we make the following contributions.

\smallskip
{\em Firstly}, to make full use of edge computing, vehicle-road collaborative infrastructures, and the emerging foundation model related technologies, we propose a new scenario that can benefit various types of smart city applications. We then mathematically model our online joint optimization of order-serving and vehicular model fine-tuning. To enhance our method's applicability in smart city contexts, our decision variables include moving decisions and data collecting decisions for {\em idle} vehicles rather than merely accepting orders, which should strike a balance in gaining high order utility and crowdsensing quality. This approach seeks to exploit the vehicles' untapped potential for both dynamic exploration of orders and service for data collection, diverging from current platforms that limit decisions to route planning for order serving. Besides, we integrate the data freshness consideration into model fine-tuning tasks. Our optimization objective is then defined as the weighted sum of the driver's total utility derived from both order serving and VCS tasks. 


\smallskip
{\em Secondly}, we propose an MARL framework that captures the interaction between vehicles and the urban environment by graph neural networks (GNN). The relationship between vehicles and environmental factors is conceptualized as a topology graph structure. Then, we use relational graph convolutional networks (R-GCN) to encode the state space and provide a global view for the agent through information propagation on the graph topology. GNN helps agents capture global information while ensuring that the input dimension of the agent's policy network is fixed. 
To further optimize the fine-tuning process, we introduce the RankTuner module, which dynamically adjusts the LoRA rank to balance fine-tuning accuracy and efficiency.
This ensures that agents can adaptively approach optimal performance in unknown and evolving environments. Diverse reward functions tailored to specific decisions are designed within the framework. 

\smallskip
{\em Finally}, we design a simulator to capture the environmental dynamics and complex vehicles' activities. We then evaluate our algorithm using the real-world New York Taxi dataset to simulate ride order generation and conduct extensive experiments of UFM fine-tuning for various tasks. Through comparative analysis with baselines, our method demonstrates superior performance in both order-serving and model fine-tuning, as well as the aggregated utility.


\smallskip
We organize the rest of the paper as follows. The related work will be introduced in Section II. Then we present our formulation in Section III, as well as our MARL solution method in Section IV. In Section V we describe the experimental setup and show the experimental results. In Section VI, we summarize and conclude the full paper.

\section{Related Work}
In this section, we provide insightful existing works in five related categories of research areas. We also highlight the key differences between these works and our study, in each paragraph.


\smallskip
\noindent\textbf{Vehicle Crowdsensing (VCS).}
In VCS systems, vehicles move within defined ranges and collect data for performing VCS tasks, e.g., machine learning driven traffic monitoring. Various studies consider optimizing different metrics for these tasks, such as the volume of data collected, the geographic fairness of the collection, or the sensing coverage. For example, Fan \emph{et al.}~\cite{INFO21} formulate the VCS problem as a stochastic dynamic program and achieve fine-grained spatio-temporal sensing coverage at the minimum long-term cost. Other works~\cite{INFO23Privacy}, \cite{INFO22Privacy}, \cite{TITS2020Privacy} focus on protecting the privacy of the vehicles participating in the crowdsensing tasks. In~\cite{vehiclecrowdsensingdeepapproach}, researchers explore the optimal selection of mobile vehicles using a deep learning-based offline algorithm that predicts vehicle mobility. Recently, to overcome the constraint of ground-only operations, unmanned aerial vehicles (UAVs) have been incorporated for data collection, collaborating with ground vehicles. Yu \emph{et al.}~\cite{ICDE23} maximize the amount of collected data and minimize energy consumption through the cooperation between unmanned ground vehicles (UGVs) and UAVs. 
Furthermore, the EHTA framework\cite{lu2024ehta} introduces an environmentally-aware heterogeneous task allocation mechanism to ensure fairness and efficiency in vehicular crowdsensing.
In terms of traffic monitoring, Li \emph{et al.}~\cite{li2019privacy} introduced a novel security model to define the misbehavior of malicious drivers in traffic monitoring via fog-assisted vehicular crowdsensing.
Notably, our approach diverges from these studies as we study a new scenario where ride-hailing vehicles can gain utilities in joint tasks of order serving and model fine-tuning.

\smallskip
\noindent\textbf{Model Fine-tuning in Vehicular Networks.}
Numerous studies have explored the integration of machine learning (ML) tasks within vehicular networks, demonstrating their potential to enhance intelligent transportation systems. For instance, ML algorithms have been employed for tasks such as real-time traffic prediction\cite{sun2020machine,chawla2024real}, dynamic routing optimization\cite{kumar2023dynamic}, and autonomous vehicle decision-making\cite{shi2020automated}. These studies underscore the growing reliance on vehicular networks for executing computationally intensive tasks in distributed environments. Recent works have increasingly focused on fine-tuning models for vehicular networks, particularly leveraging RSUs and edge devices for collaborative learning. GIOV \cite{xie2024giov} proposed a federated learning framework using RSUs to fine-tune models for adaptive vehicular applications, achieving enhanced performance in resource-constrained environments. Similarly, GAI-IOV \cite{xie2024gai} explored the deployment of pre-trained generative models for personalized content delivery in connected vehicles, addressing privacy concerns through local fine-tuning mechanisms. Otoum \emph{et al.}~\cite{otoum2022transfer} introduced an energy-efficient strategy for fine-tuning models across vehicular networks, focusing on minimizing latency and resource consumption during updates.In parallel, fine-tuning UFMs has proven transformative in addressing specific challenges in urban environments. For example, GeoSAM \cite{sultan2023geosam} enhances the SAM \cite{kirillov2023segment} model for mobility infrastructure segmentation by integrating automated visual prompt generation, showcasing its potential for urban planning. RingMo-SAM \cite{yan2023ringmo} customizes SAM’s prompt encoder for multi-source remote sensing segmentation, significantly improving performance with complex urban datasets. Additionally, GeoCLIP \cite{vivanco2024geoclip} fine-tunes the CLIP \cite{radford2021learning} framework to align images with geographic coordinates for global-scale geo-localization, utilizing innovative GPS encoding and hierarchical representations to enhance performance even with sparse training data.In contrast to these studies, our work introduces a novel scenario that synergistically benefits a diverse array of smart city applications. We not only mathematically formulate the online joint optimization of order-serving and vehicular model fine-tuning, capturing the intricate dynamics between vehicles and their urban environments, but also consider critical factors often overlooked. Specifically, our approach emphasizes the age of fine-tuning data, recognizing its significance in maintaining model relevance in dynamic urban contexts, and examines the impact of data volume on fine-tuning accuracy, ensuring robust performance even under constrained conditions. This holistic perspective enhances both the practicality and effectiveness of vehicular model fine-tuning in real-world applications.

\smallskip
\noindent\textbf{Vehicle Dispatching.}
Recently, with the rapid development of ride-hailing platforms, such as Didi~\cite{DidiChuxing2020}
and Uber~\cite{Uber2020}, more and more private car owners are opting to become ride-hailing drivers. In a large-scale fleet system, it makes sense to reasonably dispatch vehicles to maintain a balance between supply and demand. Traditionally, vehicle dispatching problems are commonly approached as classic combinatorial optimization problems, exemplified by the Traveling Salesman Problem (TSP) or the Vehicle Routing Problem (VRP). 
For instance, Zhang \emph{et al.}~\cite{KDD17} propose a novel combinatorial optimization model for the dispatching problem maximizing the global success rate by optimally matching drivers and riders, thus enhancing overall travel efficiency and improving the user experience.
In~\cite{zhang2021data}, a data-driven optimization method is deployed on the transportation network company's side to efficiently and effectively schedule the vehicles’cruising plan.
Yuen \emph{et al.}~\cite{WWW19} propose performing dynamic programming to help drivers find the route most likely to pick up passengers and maximize the probability of finding additional compatible customers while minimizing detours beyond a permissible threshold.
Other works may choose different optimization goals such as vehicle cruising time minimization~\cite{Ubiquitous18} or constraints such as energy consumption limits~\cite{vehicle_dispatch_modle_based},\cite{ge2023towards}. 
These works require a relatively complex mathematical model to model the distribution of orders or environmental dynamics, e.g., traffic or weather conditions, and their common focus is order serving provided to passengers. Our research distinguishes itself from theirs as our vehicles also bear the responsibility of data collection and model fine-tuning. Another difference is that, some of the above works consider route planning, but our work does not focus on specific route arrangements. Instead, as shown in Fig. \ref{fig:system overview}, our controlled decision variables are the next locations each vehicle should move to, rather than specific roads. This finer-grained dispatching, compared with planning routes, enhances the flexibility for idle vehicles to adapt with changing environments and increases the probability to maximize task performing utilities. 


\noindent\textbf{Deep Reinforcement Learning (DRL).}
DRL is a powerful technique that uses neural networks to simulate or evaluate the actions of agents that play actions in sequential timesteps through the feedback from environments and revealed states~\cite{introduce_drl}. In recent years, a series of research works applied DRL to VCS or vehicle dispatching systems~\cite{TITS22}. For instance, in \cite{INFO20}, a 
IMPALA-based~\cite{espeholt2018impala} framework is proposed for multi-task-oriented VCS, which features a centralized control and distributed execution system. Ding \emph{et al.}~\cite{INFO21VEHICLES} use graph neural networks to extract road network information and combine it with reinforcement learning to select routes for taxis participating in the VCS task. In these studies, the data distribution is established at the initial moment, and vehicles continuously collect data from the environment while in motion. However, in our scenario, vehicles are dynamically directed to arrive at data PoIs, where urban data can be generated and discarded in real time. We design novel reward and optimization metrics to incorporate data freshness as a pivotal factor influencing the utility of UFM fine-tuning task. In related areas such as vehicle dispatching, DRL is also widely used and shows superior performance. Some studies~\cite{INFO18, TITS20} regard the dispatch center as an agent responsible for making dispatch decisions for vehicles. Lin \emph{et al.}~\cite{KDD18} propose a contextual multi-agent reinforcement learning framework that can capture the complicated stochastic demand-supply variations in high-dimensional space. Sun \emph{et al.}~\cite{KDD22} treat each driver as an agent and solve joint order dispatching and driver repositioning problems by reinforcement learning. 
In SA-MADRL~\cite{liu2024multi}, each idle mobile charging station or each electric vehicle with insufficient electricity obtains the surrounding situation regarding the charging supply and charging demand through V2V communications, and then uses a Q-network trained by multi-agent deep reinforcement learning method to make the scheduling decision independently.
These DRL-based vehicle dispatching studies do not consider model fine-tuning tasks fulfilled by vehicles, and thus their problems do not have the challenge of addressing spatio-temporal heterogeneity in either data or fine-tuning tasks.

\smallskip
\noindent\textbf{Graph Neural Networks (GNNs)}
Given the dynamics and potential complex-structured items in the environments, DRL algorithms may benefit from enhanced state representation. Graph Neural Networks (GNNs) is an advantageous tool for state representation to extract the graph-structured dependencies across different entities in DRL environments. Recently, Li \emph{et al.}~\cite{liyihong} nicely integrate Heterogeneous Graph Neural Network (HAN) techniques and attention mechanisms into an MARL framework for minimizing the aggregate completion time for machine learning tasks in distributed edge cluster serving systems. You \emph{et al.}~\cite{you2024raccoon} jointly optimize the content recommendation and caching in IVI systems, and they propose a GNN for user-item prediction to capture the intricate topological relationship between different agents, outperforming the conventional MARL strategies in agent coordination. 
For instance, Pamuklu \emph{et al.}\cite{pamuklu2023heterogeneous} propose a heterogeneous GNN-RL-based approach for task offloading in multi-UAV networks, where a novel GNN is used to model the relationships between vertices via graph network chains. MAGIC\cite{niu2021multi} employs Graph Attention Networks (GATs) to model and manage the interactions between agents in agent-agent communication. Rathore \emph{et al.}\cite{rathore2023gnn} introduce a modified graph structure that incorporates both vehicle topology and reputation estimates from Roadside Units (RSUs), leveraging GNN to enhance the reinforcement learning process by enabling vehicles to fuse sensor data and peer-reported safety messages for more accurate reputation estimation.  In contrast to prior work, which focuses on static or predefined relationships in task-specific graphs, our approach introduces a dynamic state space representation that evolves with time and context. By leveraging dynamic topology graph, we model the complex interactions between vehicles, orders, PoIs, and grids within an urban environment. Our method captures the full complexity of the environment, enabling better performance across a wider range of tasks.

\section{Problem Formulation}
\subsection{System Overview}
\label{subsec:sys-overview}


\begin{table}[t]

\caption{Key Notations with Description.}
\label{table:notation}
\centering
\begin{threeparttable}
\begin{tabular}{c p{6cm}}
\toprule
Notation & Explanation \\
\midrule
{$m$, $M$, $\mathcal{M}$} & Vehicle index, number of vehicles, Vehicle set \\

\midrule
{$g$, $G$, $\mathcal{G}$} & Grid index, number of grids, Grid set\\

\midrule
{$t$, $T$, $\mathcal{T}$} & Time slot index, number of time slots, time slot set\\

\midrule
{$o$, $\mathcal{O}_t$} & Order index, Order set\\

\midrule
{$p$, $\mathcal{P}_t$} & PoI index, PoI set\\

\midrule
{$g_t^m$} & Index of grid vehicle $m$ is located\\

\midrule
{$\mathcal{N}(g)$} & Neighboring grids set of grid $g$\\

\midrule
{$\sigma(o)$} & Price of order $o$\\

\midrule
{$d\left(p\right)$} & Data volume of PoI $p$\\

\midrule
{$\eta$} & The rank of the low-rank matrices in LoRA\\

\midrule
{$\lambda_{t}^p$} & Data AoI of PoI $p$\\

\midrule
{$u_{t}^p$} & The data utility of the PoI $p$\\

\midrule
{$\textbf{x}$, $\textbf{y}$, $\textbf{z}$} & Vehicle dispatching, order accepting, data collecting\\

\midrule
{$s_{m,t}$, $a_{m,t}$, $r_{m,t}$} & State, action and reward of vehicle $m$\\

\bottomrule
\end{tabular}
\begin{tablenotes}   
        \footnotesize               
        \item[1] $\bullet$ We omit descriptions such as ``at time slot $t$".     
      \end{tablenotes}           
    \end{threeparttable}

\end{table}

In this section, we provide a detailed description of our scenario. We consider a vehicular network, consisting of a large number of moving ride-hailing vehicles that are managed by a cloud platform (such as DiDi or Urber) within a certain geographical range and can interact with RSUs equipped with edge servers. Beyond the routine operations of picking up and dropping off passengers, each vehicle actively engages in urban sensing tasks. Leveraging the installed professional sensors and smartphones, the vehicles can collect data from PoIs distributed across the designated area. We define $\mathcal{M} \triangleq \left\{m | m = 1,2,\cdots, M\right\}$ to represent the set of vehicles in the system. The activity range of each vehicle is limited to the target area. Similar to many prior studies \cite{KDD18, KDD22}, we discretize the target area into $G$ grids, represented by set $\mathcal{G} \triangleq \left\{g|g = 1, 2, \cdots, G \right\}$. Similarly, the time horizon is also divided into several discrete time slots, represented by $\mathcal{T} \triangleq \left\{t| t = 1, 2, \cdots, T \right\}$. The set of orders and the set of PoIs at time slot $t$ are $\mathcal{O}_t$ and $\mathcal{P}_t$, respectively. At time slot \( t \), the set of orders is denoted as \( \mathcal{O}_t \), and the set of Points of Interest (PoIs) is represented as \( \mathcal{P}_t \). 
To account for task-specific requirements, \( \mathcal{P}_t \) can be decomposed into subsets, where each subset corresponds to a distinct type of data PoI associated with a particular task. Formally, this decomposition can be expressed as:  
\begin{align}
    \mathcal{P}_t = \bigcup_{k \in \mathcal{K}} \mathcal{P}_{t,k}, \quad \mathcal{P}_{t,k} \cap \mathcal{P}_{t,j} = \emptyset, \, \forall k \neq j,
\end{align}
where \( \mathcal{K} \) is the set of task types, and \( \mathcal{P}_{t,k} \) denotes the subset of PoIs related to task \( k \). For instance, \( \mathcal{P}_{t,1} \) represent PoIs contributing to Vision Transformers (ViT)\cite{dosovitskiy2020image} based tasks such as image classification, while \( \mathcal{P}_{t,2} \) corresponds to PoIs relevant to SAM-based tasks like image segmentation. 
Since the number of orders and PoIs change at each time slot, the size of $\mathcal{O}_t$ and $\mathcal{P}_t$ can be distinct across $t$. We consider vehicles that currently not serving orders or collecting data as available vehicles. For grid $g$, we use $\mathcal{O}_t^g$, $\mathcal{M}_t^g$ and $\mathcal{P}_t^g$ respectively to represent the set of orders, the set of available vehicles, and the set of PoIs in grid $g$ at time slot $t$. The index of the grid (interchangeable with location in this paper) where vehicle $m$ is located at time slot $t$ is $g_t^m$, where we have $g_t^m \in \mathcal{G}$. To maximize the effectiveness of order-serving and data collection, each available vehicle $m$ needs to decide whether to accept an order $o \in \mathcal{O}_t^{g_t^m}$, collect data from a PoI $p \in \mathcal{P}_t^{g_t^m}$, or travel to another grid. Once an available vehicle 
$m$ accepts an order or collects data from a PoI, it becomes unavailable and will revert to being available again after completing passenger drop-off or data collection. For convenience, some of the important notations used in the paper are listed in Table \ref{table:notation}.

\subsection{Primer on Model Fine-tuning with LoRA}
\label{ssec:primer-lora}
Low-Rank Adaptation (LoRA) is a powerful technique designed to efficiently fine-tune large pre-trained models, without the need to retrain all model parameters. Traditional fine-tuning methods often involve updating the entire model’s weights, which can be computationally expensive and resource-intensive, especially with massive models. LoRA addresses this by observing that the difference between the pre-trained weights and the fine-tuned weights often lies in a low-dimensional subspace. Thus, LoRA introduces a low-rank approximation to model these changes, significantly reducing the number of parameters that need to be updated. The benefits of LoRA include a substantial reduction in memory and computational costs, as well as the ability to adapt large models to new tasks with minimal overhead.
Formally, LoRA modifies the weight matrix $W_0 \in \mathbb{R}^{d \times k}$ of a pre-trained model by introducing a low-rank update, represented as $W_0 + \Delta W = W_0 + BA$, where $B \in \mathbb{R}^{d \times \eta}$ and $A \in \mathbb{R}^{\eta \times k}$, with $\eta \ll \min(d, k)$. The matrices $B$ and $A$ are the only trainable parameters, while the original weights $W_0$ are frozen.  The model input is denoted as $x$, and the output is represented as $h$. The forward pass incorporating the LoRA module is given by:
\begin{align}
h = W_0 x + \gamma \eta BA x,
\end{align}
where \( \gamma \) is a scaling factor, and \( \eta \) is the rank of the low-rank matrices. This formulation shows how LoRA modifies the original model by adding a low-rank update in parallel to the pre-trained weights. The rank \( \eta \) controls the capacity of the low-rank adaptation, allowing for efficient fine-tuning with minimal changes to the original model. Additionally, LoRA preserves computational efficiency during inference, making it particularly useful for deployment in resource-constrained environments.

\subsection{QoS and Optimization Modeling for Vehicular Joint Tasks}
\label{ssec:optimization-qos}

Subsequently, we define the Accumulated Driver Income (ADI), the Accumulated Data Utility (ADU), and the QoS. Following this, we give the mathematical form of the problem and the optimization goal.

\noindent\textbf{Definition 1. (ADI)} We use $\mathcal{O}_t^{'g}$ to denote the set of orders accepted by vehicles in grid $g$ at time slot $t$. For order $o$ within the set $\mathcal{O}_t^{'g}$, we denote by $\sigma(o)$ the price of order $o$. ADI represents the total income of all drivers over all time slots, so the expression for ADI is as follows: 
\begin{align}
ADI 
    = \sum\limits^{T}_{t=1} \sum\limits^{G}_{g=1}\sum\limits^{}_{o \in \mathcal{O}_t^{'g}} \sigma(o).
\end{align}

There are several PoIs distributed within each grid, and vehicles can collect data from these PoIs. Each PoI $p \in \mathcal{P}_t$  is associated with a certain volume of data $d(p)$ that needs to be collected. To maintain data integrity and prevent excessive bandwidth usage, we assume that an available vehicle can collect data from only one PoI at a time, and the collection process continues until the pending data volume at the PoI is reduced to 0.

\noindent\textbf{Definition 2. (ADU)} We define ADU as the sum of the data utility collected by all vehicles in all time slots. The expression for ADU is defined as:
\begin{align}
ADU =  \sum\limits^{T}_{t=1} \sum\limits^{G}_{g=1} \sum\limits^{}_{p \in {\mathcal{P}}_t^{'g}} u_t^p,
\end{align}
where $\mathcal{P}_t^{'g}$ is the set of PoIs collected in grid $g$ at time slot $t$. We use ADU to indicate the quality of VCS in the following sections.

\begin{figure}[t]
\hspace{-1cm}  
\subfigure[Accuracy as a function of data freshness or AOI for two different tasks.]{
\label{fig: sub.1 aoi}
\includegraphics[width=0.5\textwidth]{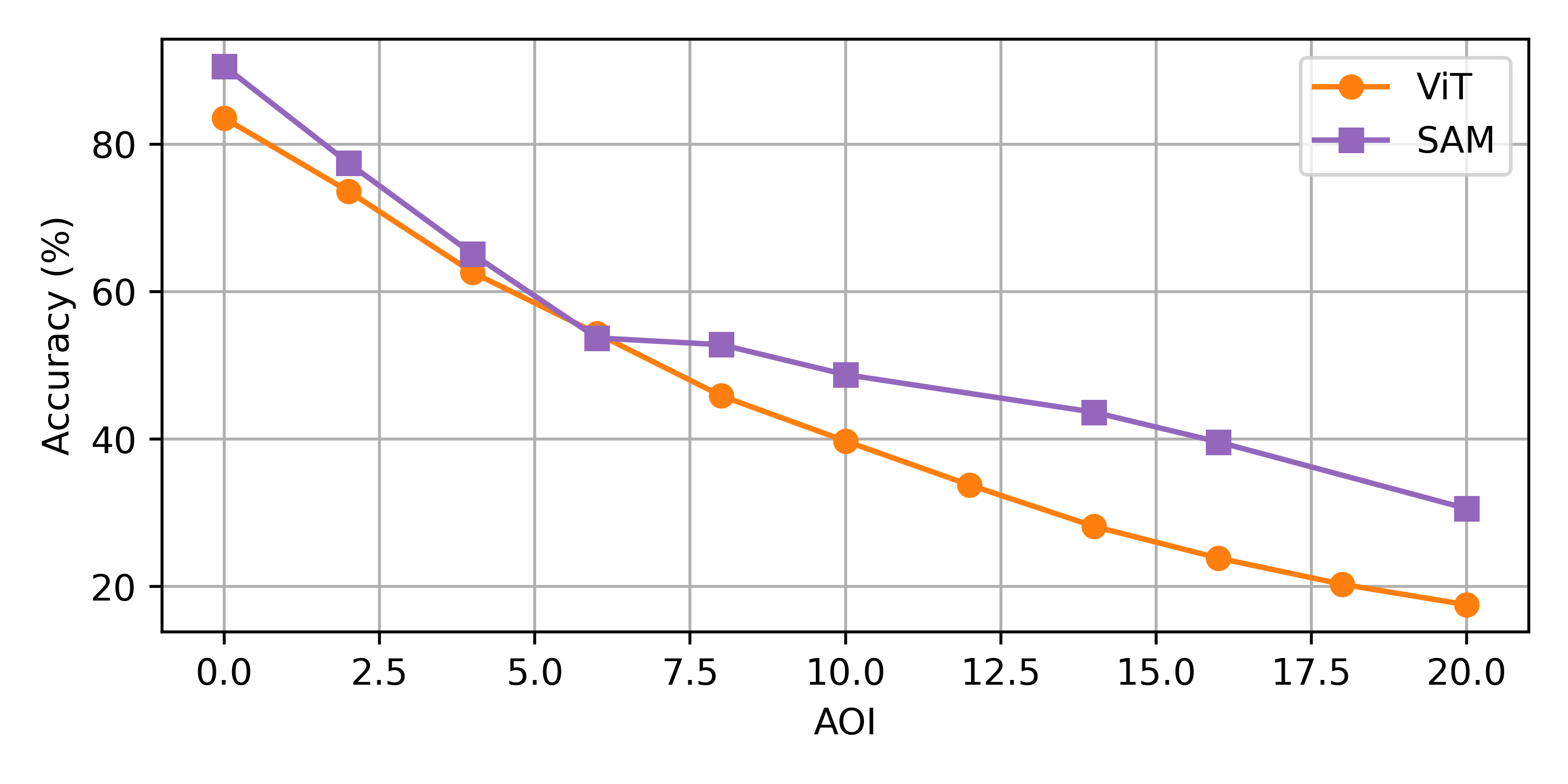}}
\subfigure[Accuracy as a function of data volume for the same tasks.]{
\label{fig: sub.2 data_volume}
\hspace{-0.4cm}  
\includegraphics[width=0.5\textwidth]{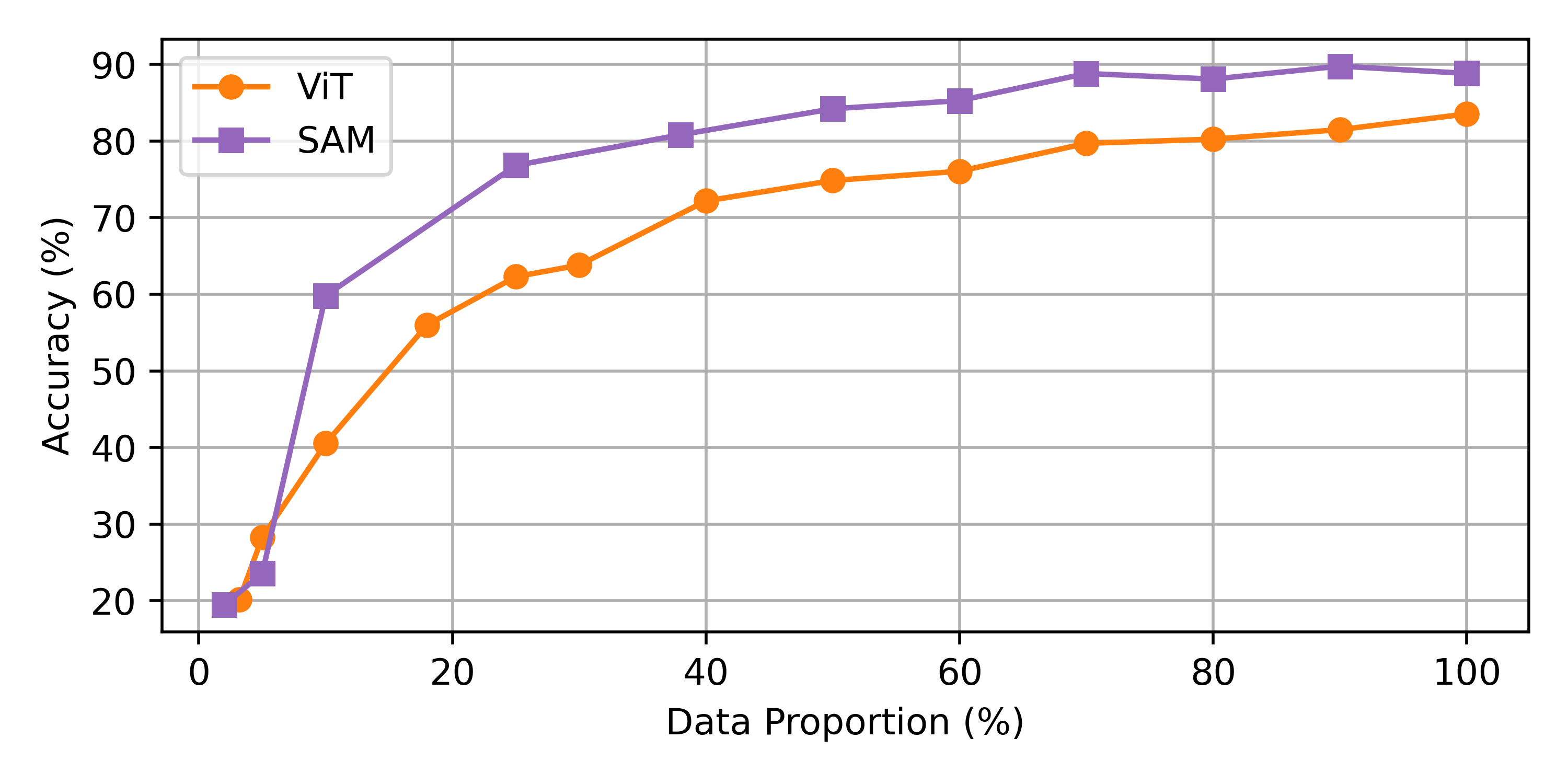}}
\caption{The impact of data freshness and data volume on the fine-tuning accuracy of different tasks under different UFMs.}
\label{fig: impact_on_fine-tune}
\end{figure}

\noindent{\bf Freshness and quantity based data utility.} Many existing studies tend to overlook the importance of data freshness in model fine-tuning, yet we explicitly take into account the effects of using various freshness degrees of collected data to fine-tune the foundation model using LoRA techniques, and we focus on the inference accuracy of such fine-tuned models varying data freshness. Intuitively, the freshness of data should hold significant relevance in numerous real-time VCS tasks \cite{the_reason_why_data_freshness_is_important}. For instance, in traffic monitoring, the most recent traffic data is more beneficial for intelligent transportation systems \cite{a_example_for_the_reason_why_data_freshness_is_important}. But these works are not foundation fine-tuning tasks and the task performance has different metrics from ours. 
To verify this intuition, we conducted a series of experiments on various large model fine-tuning tasks, including LoRA-based fine-tuning for image classification using ViT and image segmentation using SAM. As Fig.~\ref{fig: impact_on_fine-tune} show, the relationship between fine-tuning accuracy and data freshness exhibits a complex and often unpredictable pattern. Specifically, accuracy tends to follow a concave or even linear decay as data freshness decreases, though the exact functional form is task-dependent and difficult to predict. This underscores the challenge of modeling data freshness in real-world applications. To address this uncertainty, we propose leveraging DRL to learn optimal fine-tuning strategies for varying data freshness. Additionally, our experiments reveal that accuracy improves with the volume of data, but the relationship is not linear. As the amount of data increases, the fine-tuning accuracy shows a logarithmic growth, and beyond a certain threshold, further increases in data volume yield diminishing returns. This phenomenon is consistent with the scaling laws for neural language models\cite{kaplan2020scaling}, which highlights the diminishing impact of additional training data after reaching a critical dataset size. These results emphasize the importance of considering both data freshness and quantity when fine-tuning UFMs, and highlight the potential of DRL in optimizing these factors for improved task performance.

\smallskip
\noindent{\bf Spatio-temporal heterogeneity and task differences.} 
Data collection for fine-tuning tasks exhibits significant spatio-temporal heterogeneity and task-specific variations. Spatially, different PoIs are associated with distinct distributions of collected datasets, influenced by factors such as regional activity patterns, environmental features, and task-specific requirements. These datasets are further tied to varying types of tasks and corresponding base models, such as image classification using ViT or image segmentation with SAM. This spatial diversity necessitates careful consideration of the data's regional context when fine-tuning models. Temporally, the heterogeneity arises from the varying arrival times of vehicles at PoIs. As vehicles may visit a PoI at different times, the freshness of the collected data naturally declines over time, as discussed in the previous section. This temporal decay in data freshness introduces additional complexity to tasks that rely on timely and accurate data for model fine-tuning. Balancing these temporal factors is critical for maintaining the relevance and utility of the collected data.
For better comprehension, we consider the price of an order as the utility of the order. The utility of an order remains constant from the time of its creation until its expiration, during which the utility that a driver can obtain by accepting the order is equivalent to the order's price. If the order expires, its utility abruptly drops to 0. The variation trends in the utility of orders and PoIs are distinct, implying that accepting orders or collecting data at suitable times and locations may potentially minimize the impact between the profit generated from passenger transportation and the utility derived from data collection. 

\noindent\textbf{Definition 3. (QoS)} To finally capture the overall performance of order-serving and fine-tuning tasks, we propose a QoS function to evaluate the overall utility of all the vehicles in the network that accomplish both order serving and model fine-tuning tasks, i.e.,
\begin{align}
QoS = \alpha ADI + \beta ADU, \label{equ: qos def}
\end{align}
where $\alpha$ and $\beta$ are importance factors that balance the contributions of ADI and ADU, respectively. These weights are determined collaboratively by stakeholders, such as government entities or organizations overseeing vehicle-sensing coordination tasks and ride-hailing companies responsible for passenger service. By setting these hyperparameters in advance, the system adapts to operational priorities and ensures alignment with strategic objectives.

In our settings, the operations that each available vehicle can perform are denoted as \textbf{x}, \textbf{y}, and \textbf{z}, corresponding to dispatching, order acceptance, and data collection, respectively. The dispatching decision \textbf{x} exerts a certain impact on both \textbf{y} and \textbf{z}, and has a long-term impact on QoS. Specifically, in real-world applications, the distribution of orders in the city is non-uniform, and a sequence of dispatch decisions for a vehicle will allocate the vehicle to grids with varying order quantities. The vehicle assigned to grids with a lower supply-demand ratio has more opportunity to match an order successfully. Although the distribution of PoIs is often related to the type of VCS task, in some cases the distribution of PoIs is not uniform and the data utility of different PoIs in different grids is also different. We express the mathematical form of our optimization problem as follows.
\begin{align}
\underset{\mathbf{\textbf{x},\textbf{y},\textbf{z}}}{\text{Max}}
 &\hspace{3mm}
 {QoS(\textbf{x}, \textbf{y}, \textbf{z})}& \label{max obj}\\
 \text{S.t.} 
 &\quad x_t^{m, g} \in \left\{0,1\right\}, \forall m \in \mathcal{M}, \ t \in \mathcal{T}, \ g \in \mathcal{N}(g_t^m) \label{cst: x binary}\\
 & \quad y_t^{m} \in \left\{0,1\right\}, \forall m \in \mathcal{M}, \ t \in \mathcal{T} \label{cst: y binary}\\
 & \quad z_t^{m} \in \left\{0,1\right\}, \forall m \in \mathcal{M}, \ t \in \mathcal{T} \label{cst: z binary}\\
 & 0 \leq \sum\limits^{}_{g \in {\mathcal{N}(g_t^m)} } x_t^{m, g} \leq 1, \forall m \in \mathcal{M}, \ t \in \mathcal{T} \label{cst: one x} \\
 & \sum\limits^{}_{g \in {\mathcal{N}(g_t^m)} } x_t^{m, g} + y_t^m + z_t^m = 1, \forall m \in \mathcal{M}, \ t \in \mathcal{T} \label{cst: one choice}
\end{align}

Here, constraints (\ref{cst: x binary}), (\ref{cst: y binary}), and (\ref{cst: z binary}) denote that vehicle dispatching decision $x_t^{m,g}$, order-accepting decision $y_t^{m}$, and data-collecting decision $z_t^m$ are binary. In constraint (\ref{cst: x binary}), $g_t^m$ represents the grid where vehicle $m$ is currently located, and $\mathcal{N}(g)$ represents the set of grids adjacent to grid $g$ (including grid $g$ itself). For each vehicle, $x_t^{m,g}$ is only positive (= 1) when vehicle $m$ is dispatched to grid $g$ at time slot $t$, otherwise it is 0 (constraint (\ref{cst: x binary})). When $x_t^{m,g_t^m} = 1$, it means that vehicle $m$ decides to stay in its current grid $g_t^m$. $y_t^{m}$ is 1 only when $m$ decide to accept an order at time slot $t$, otherwise 0 (constraint (\ref{cst: y binary})). Similarly, $z_t^{m}$ is 1 only when $m$ decides to collect data from a PoI at time slot $t$, otherwise 0 (constraint (\ref{cst: z binary})).
Inequality (\ref{cst: one x}) denotes that vehicle $m$ has only one dispatch destination at a time slot. Equation (\ref{cst: one choice}) denotes that at each time slot, vehicles choose one of three choices: traveling to the dispatching destination $g \in \mathcal{N}(g_t^m)$, accepting an order, and collecting data. Note that the vehicles we mention in this section refer to the ones that are available at the current time slot.

Our mathematical models (\ref{max obj})-(\ref{cst: one choice}) describe the joint ride-hailing vehicles dispatching and crowdsensing scenarios. This is an online and sequential decision-making problem where decisions need to be made in real-time based on available information. Reinforcement learning emerges as a superior approach for addressing such problems. Our method is introduced in the next section.

\section{Proposed Solution: GNN-enhanced MARL}

In this section, we provide a comprehensive description of the algorithms designed to address the challenges posed by the joint scenarios of ride-hailing vehicle dispatching and crowdsensing. Our algorithm leverages MARL and integrates with other technologies to optimize the QoS as defined in (\ref{max obj}). 

\subsection{Algorithm Overview}
\begin{figure*}[t]
\centerline{\includegraphics[width=1.0\linewidth]{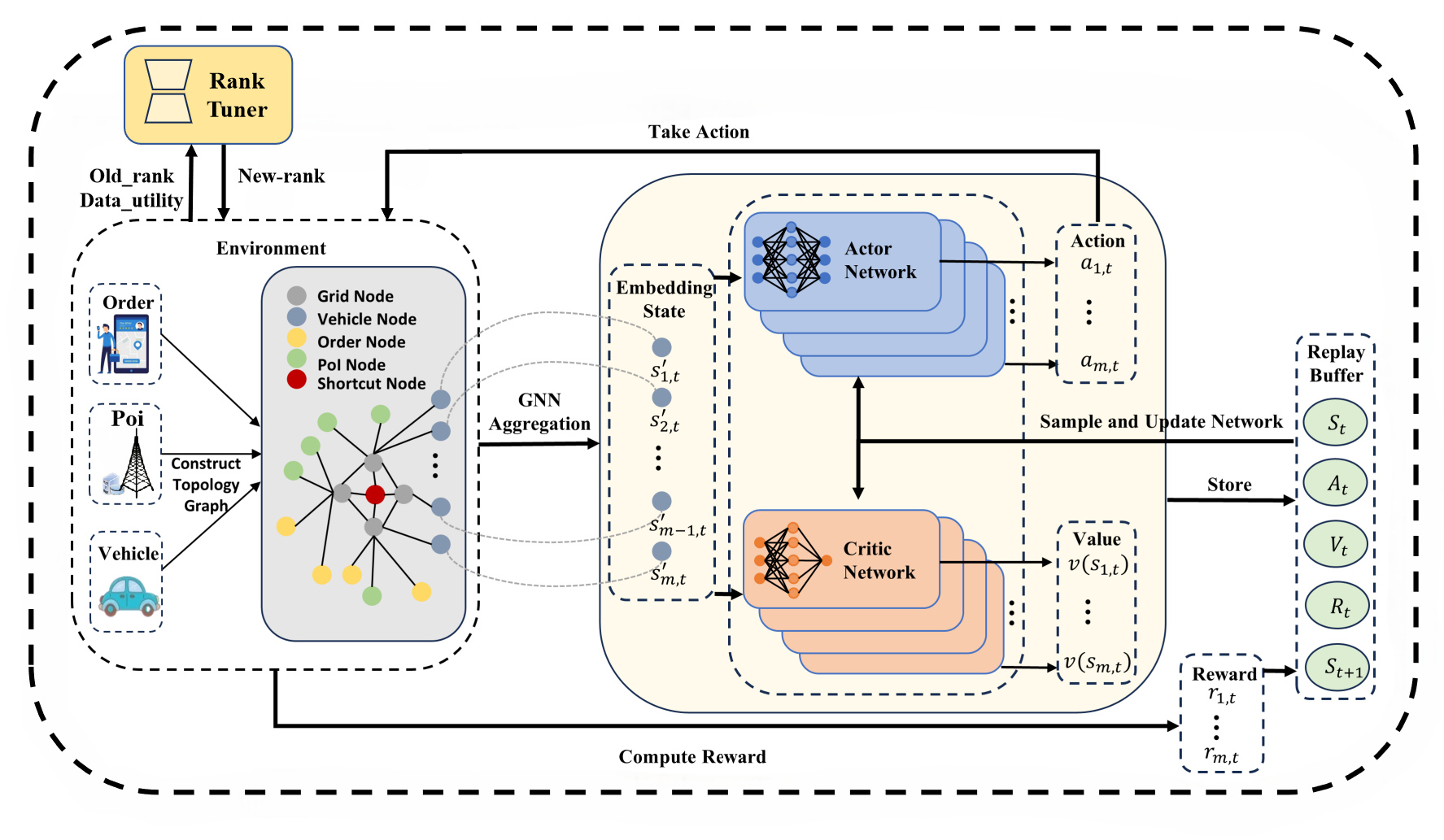}}
\caption{GNN-based MARL framework. It includes the environment, a GNN embedding module for processing raw state information, an actor-critic module for decision-making, a replay buffer for experience storage, and the RankTuner module for dynamically adjusting LoRA ranks to balance fine-tuning accuracy and efficiency. These components work together to enable agents to make independent, informed decisions while optimizing their actions based on the dynamic environment.
}
\label{fig: module intro}
\end{figure*}

Due to the complexity of joint optimizing decision variables for fleets in urban environments, we employ a decentralized optimization framework. For each time slot, decision variables $\textbf{x}$, $\textbf{y}$ and $\textbf{z}$ can be independently determined by each available vehicle. Optimizing the QoS becomes feasible if the framework can learn the statistical distribution of orders and PoIs while gaining insights into their time sensitivity. The utilization of a distributed optimization framework effectively mitigates the challenge of the decision space rapidly expanding with the size of the vehicle set. The demand for online optimization and distributed decision-making motivates us to use MARL, renowned for its excellent performance in long-term and coupled decision-making \cite{why_use_MARL}. We illustrate the structure of our framework in Fig. \ref{fig: module intro}. Our system is based on the actor-critic MARL algorithm, with each agent making independent decisions. Additionally, we incorporate GNN for raw state embedding. 
Furthermore, the framework integrates the RankTuner module, enabling dynamic adjustment of LoRA ranks to balance fine-tuning accuracy and efficiency. 
In the subsequent sections, we provide a concise overview of key settings such as states, actions, and rewards, followed by an explanation of GNN-based state embedding. 
\subsection{MARL Statement}
Formally, we model the joint optimization problem as a Markov game, which is represented by a
tuple $(\mathcal{S}, \mathcal{A}, \mathcal{P}, \mathcal{R})$, where $\mathcal{S}$, $\mathcal{A}$, $\mathcal{R}$ and $\mathcal{P}$ are the set of states, actions, rewards, and state transition probability. We define the important components in the MARL framework as follows.



\noindent\textbf{Agent.} We consider a vehicle as an agent. As our objective is to optimize the overall income and data utility of all vehicles, each vehicle can be considered as a homogeneous agent performing cooperative tasks. We still use $\mathcal{M}$ to represent the set of agents.

\noindent\textbf{State space.} Intuitively, a global environmental state should encompass factors such as the distribution of drivers, the distribution of orders, the distribution of PoIs, the current time slot $t$, the generation time and estimated travel time for each order, and the data volume and AoI for each PoI. At the beginning of each time slot $t$, each vehicle $m$ gets a local state correlated with the global environment state $s_t \in \mathcal{S}$, which can be written as ${s_{m,t}}$. The challenge arises in determining whether to aggregate all the relevant factors into a large state space for every agent or to partition the state space into subspaces for each agent. Both approaches prove to be inefficient. Moreover, the continuous changes in the number of orders and PoIs lead to a dynamic change in the state space dimension over time. Selecting a fixed dimension for the state space becomes impractical, posing challenges for implementing the MARL algorithm. To tackle these challenges, we leverage Relational Graph Convolutional Networks (R-GCN) \cite{R-GCN} to encode the features of each agent. The output of R-GCN, denoted as $s^{\prime}_{m,t}$, serves as the embedding state of each agent, integrating the raw state and reducing the raw state dimension to a fixed dimension. The concrete representation of the state space and the detailed process of state embedding will be presented in the subsequent subsection.



\noindent\textbf{Action space.}
At every time slot, each available agent $m$ takes an action $a_{m,t}$ according to its policy after getting the embedding state $s^{\prime}_{m,t}$. The action $a_{m,t}$ indicates whether the agent should be dispatched to a neighboring grid, remain in the current grid and whether it should accept an order or collect data from a PoI. We denote all agents' joint action as $a_t = \left\{a_{1,t}, a_{2,t}, ..., a_{M,t}\right\}$.

\noindent\textbf{State transition probability.}
Based on the environmental state $s_t$ and the joint action $a_t$ of all agents, the environmental state will transit to the next state $s_{t+1}$ with probability $p(s_{t+1}|s_t, a_t)$.

\noindent\textbf{Reward.}
We calculate an immediate reward for each available agent according to its action. Given the three distinct action types, we formulate distinct reward functions corresponding to each action type.

\begin{itemize}
\item[$\bullet$] If dispatching agent $m$ to a neighboring grid at time slot $t$, we calculate the immediate reward of $m$ as: 
\begin{align}
    r_{m,t} = 0.
\end{align}
In this paper, remaining in the current grid is conceptualized as a special form of dispatching. We assign a reward of 0 for dispatching since dispatching doesn't directly yield rewards, although it can influence subsequent actions. Simultaneously, this is implemented to discourage agents from repeatedly dispatching between specific grids to gain rewards. Despite the reward value being 0 for dispatching, our expectation is that, through MARL, agents can still learn the influence of dispatching on order-serving and data collection, enabling a more effective dispatching policy.
\end{itemize}

\begin{itemize}
\item[$\bullet$] If agent $m$ decides to accept an order at time slot $t$, the reward of $m$ is written as: 
\begin{align}
& r_{m,t} = \alpha \cdot \sigma(o_t^m),
\end{align}
where $o_t^m$ represents the order that be accepted by $m$, and $\sigma(o_t^m)$ represents the price of $o_t^m$. Here, $\alpha$ is a weight, which is the same as in expression (\ref{equ: qos def}).
\end{itemize}

\begin{itemize}
\item[$\bullet$] 
If agent $m$ decides to collect data from a PoI at time slot $t$, the reward of $m$ is written as: 
\begin{align}
& r_{m,t} = \beta u_t^{p_t^m} = \beta f_k(d_t^{p_{t,k}^m},{\lambda}_t^{{p_{t,k}^m}}).
\end{align}
Here, $r_{m,t}$ equals the utility of the data collected by $m$, and $p_t^m$ is the PoI collected by agent $m$ at time slot $t$ with task $k$. $\beta$ is also a weight like $\alpha$. The function \( f_k(d, \lambda) \) encapsulates the fine-tuning accuracy under the combined influence of two key factors: \( d \), the amount of collected data, and \( \lambda \), the degree of data freshness (also referred to as AoI). The subscript \( k \) indicates the task-specific nature of the function, as different tasks may exhibit unique dependencies on data quantity and freshness. This function quantifies how the accuracy of the fine-tuned model varies based on these two variables. As both data quantity and freshness directly impact the utility function, the interplay between these variables determines the model’s fine-tuning performance, with \( f(d, \lambda) \) capturing this dependency in a nuanced and task-specific manner.
\end{itemize}


\subsection{State Embedding}
Reviewing the structure and influencing factors of our optimization problem, the arrival of ride orders and the generation of PoIs follow a prior but unknown distribution. To ensure adaptability to the dynamic and highly stochastic environment, each agent requires a policy network with high generalization, utilizing ample state information to formulate decisions. These decisions aim to optimize the overall QoS throughout numerous time slots from a long-term perspective. Therefore, continuous monitoring of the states of vehicles, orders, and PoIs is crucial for each agent. First, we define the features of raw state for the agent $m$, namely $s_{m,t} = \left\{ \mathcal{I}_{m,t}, \mathcal{U}_{m,t}, v_{m,t}, h_{m,t} \right\}$, serving as the input of RGCN and is formalized as follows.

\begin{itemize}
\item[$\bullet$] 
\textbf{Order feature set $\mathcal{I}_{m,t}$} contains some entities $i_{m,t} \in \mathcal{I}_{m,t}$, each of which represents the states of a order. Specifically, $i_{m,t}$ is the concatenation of the price, generation time, estimated travel time, origin, and destination of an order within the grid where $m$ is located. The feature set $\mathcal{I}_{m,t}$ not only provides information on the number and price of orders within the grid but also indicates the impact of orders on vehicle distribution due to their different destinations and travel times.
\end{itemize}

\begin{itemize}
\item[$\bullet$] 
\textbf{PoI feature set ${\mathcal{U}}_{m,t}$} is a set contains entities $u_{m,t} \in \mathcal{U}_{m,t}$. Each element $u_{m,t}$ represents the data volume, AoI, and generation location of a PoI within the grid where $m$ is located. By capturing the data volume and AoI of each PoI, the set ${\mathcal{P}}_{m,t}$ effectively characterizes the present data utility associated with individual PoIs.
\end{itemize}

\begin{itemize}
\item[$\bullet$] 
\textbf{Vehicle feature vector $v_{m,t}$} represents the location, working states (serving orders, collecting data or idle), current time slot, and index of agent $m$ (using one-hot encoding). 
\end{itemize}

\begin{itemize}
\item[$\bullet$] 
\textbf{Grid feature vector $h_{m,t}$} contains the number of orders, available vehicles, and PoIs in the current grid where the agent $m$ is located. To distinguish different grids, $h_{m,t}$ includes the grid index encoded by one-hot. Grid feature reflects the distribution of orders, vehicles, and PoIs in the environment.
\end{itemize}

In our MARL framework, if each agent $m$ can only observe the local features $\left\{ \mathcal{I}_{m,t}, \mathcal{U}_{m,t}, v_{m,t}, h_{m,t}\right\}$, there is an increased susceptibility to becoming ensnared in local optima \cite{liyihong}. A global representation of features is thus needed. However, a straightforward concatenation of all $s_{m,t}$ across all agents into a global state would result in a rapid expansion of the agent's input dimension, potentially compromising algorithm performance. In addition, since the size of the order set and PoI set are not fixed, we cannot specify a fixed input dimension for the agent. We try to use GNN to solve these problems. In our approach, we embed graph neural networks into the agent's policy network, enhancing the fusion and interaction between the agent's state and the environment's state. Below we define the topology graph used by GNN.

\begin{figure}[t]
\centering  
\subfigure[Vehicles, orders, and PoIs in grids $g_1$, $g_2$, $g_3$ and $g_4$.]{
\label{fig: sub.1 distribution}
\includegraphics[width=0.3\textwidth]{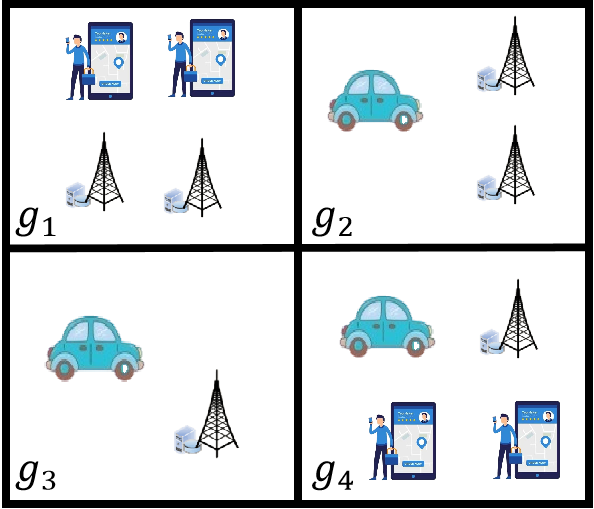}}
\subfigure[Topology graph based on (a).]{
\label{fig: sub.2 topo}
\includegraphics[width=0.32\textwidth]{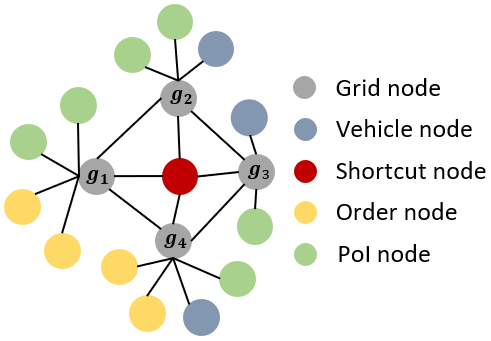}}
\caption{An example of constructing a topology graph.}
\label{fig: example of topo}
\end{figure}

\noindent\textbf{Definition 4. (Topology Graph.)} Due to the advantages of GNN in topology-based node information transferring and information processing, we represent vehicles, orders, PoIs, and grids as nodes within a topology graph. The topology graph is used to describe the relationship between urban states and vehicles. Our topology graph is $Gr(\mathcal{N}_t,\mathcal{E}_t)$, where the node set $\mathcal{N}_t$ consists of order nodes $\mathcal{O}_t$, PoI nodes $\mathcal{P}_t$, grid nodes $\mathcal{G}$, vehicle nodes $\mathcal{M}$, and a shortcut node. We abuse the notation of the order, PoI, grid, and vehicle indices to denote the corresponding nodes as well. $\mathcal{E}_t$ is the set of edges formed by connecting nodes, and its definition is as follows. As shown in Fig. \ref{fig: example of topo}, a vehicle node $m \in \mathcal{M}$ connects with a grid node $g \in \mathcal{G}$ only if vehicle $m$ is within the range of gird $g$. If the generation location of order $o \in \mathcal{O}_t$ is grid $g$, node $o$ is connected to node $g$. Similarly, PoI nodes are interconnected with grid nodes using a similar criterion. Grid node $g$ is only connected to the grid nodes which are its neighboring grid. Specifically, a shortcut node is introduced, linked to each grid node, expediting the information propagation within the graph neural network. Through the connection of grid nodes, we can capture the connectivity and topological structure between grids. Meanwhile, GNN facilitates each vehicle node's awareness of the features associated with orders and PoIs.

The topology graph contains multiple types of nodes, so we need to distinguish between different types of nodes. Since the R-GCN model can process topology graphs with different types of nodes and generate embedding information for the nodes, we use R-GCN to generate the state representation for agents. 
The information propagation is that each node $n \in \mathcal{N}_t$ passes the features as messages to its neighboring node and $n$ aggregates the features which are from its neighbors. The aggregation method and update steps of our R-GCN follow the steps in \cite{R-GCN}. The propagation model of our R-GCN is described as:
\begin{align}
    & {Gr}_{(0)}^{shortcut} = \textbf{0} \\
    & {Gr}_{(0)}^{\mathcal{M}} = \cup_{m = 1}^M v_{m,t}  \\
    & {Gr}_{(0)}^{\mathcal{G}} = \cup_{m = 1}^M h_{m,t}  \\
    & {Gr}_{(0)}^{\mathcal{O}} = \cup_{m = 1}^M \mathcal{I}_{m,t} \\
    & {Gr}_{(0)}^{\mathcal{P}} = \cup_{m = 1}^M \mathcal{U}_{m,t} 
\end{align}

The initial input ${Gr}_{(0)}$ is written as: 
\begin{align}
    {Gr}_{(0)} = \left\{ {Gr}_{(0)}^{shortcut}, {Gr}_{(0)}^{\mathcal{M}}, {Gr}_{(0)}^{\mathcal{G}}, {Gr}_{(0)}^{\mathcal{O}},{Gr}_{(0)}^{\mathcal{P}} \right \}.
\end{align} 

The node embedding is propagated in each layer $l$, i.e., ${Gr_{(l)}} = f({Gr_{(l-1)}})$, where $f(\cdot)$ represents the graph convolution network aggregating the features of each node with its neighbors. After $L$ layers of graph message passing, we get the final graph embedding $Gr_{(L-1)}$. We then map $Gr^{\mathcal{M}}_{(L- 1)}$ to the corresponding agents as the input of their actor networks.
\begin{table}[ht]
\centering
\caption{Impact of Rank on Fine-Tuning Accuracy and Time}
\resizebox{\linewidth}{!}{
\begin{tabular}{|c|c|c|}
\hline
\textbf{Rank} & \textbf{Fine-Tune Time (Normalized)} & \textbf{Accuracy Factor} \\
\hline
1 & 1.00  & 0.70     \\
2 & 1.45  & 0.76     \\
3 & 3.71  & 0.98     \\
4 & 5.05  & 0.99     \\
5 & 6.00  & 0.99365  \\
6 & 5.31  & 1.00     \\
\hline
\end{tabular}
}
\label{tab:rank}
\end{table}
\subsection{Heuristic-based Rank Selection Integrated with MARL}
Since LoRA is employed for task fine-tuning, selecting an appropriate rank is crucial. According to relevant studies~\cite{bai2024federated}, a larger rank generally leads to better fine-tuning accuracy but at the cost of increased fine-tuning time. An interesting and important trade-off here is that if choosing a larger rank greedily for obtaining a higher accuracy per task, the longer fine-tuning time will very likely reduce the expected total number of fine-tuning tasks that a vehicle can accomplish in the entire time span. Therefore, the best rank of a fine-tuning adaptor should be carefully chosen to achieve the total utility. 

Our experiments, as summarized in Table~\ref{tab:rank}, show that the choice of rank significantly impacts both accuracy (in terms of accuracy discounts compared to the best accuracy) and fine-tuning time across different tasks, such as image classification and image segmentation. To address this, we propose the RankTuner, a dynamic adjustment mechanism designed to determine the most suitable rank for fine-tuning when the optimal rank is unknown. The RankTuner operates as follows: Initially, a rank is randomly selected as the benchmark. During each iteration, the algorithm compares the current ADU to the previous round. If the ADU improves, the algorithm maintains the current direction (increasing or decreasing the rank) to further explore better settings. If the ADU decreases, the algorithm reverts to the previous rank and switches the direction. The rank is constrained within a predefined allowable range to ensure feasibility. The pseudocode for the RankTuner is presented in Algorithm~\ref{alg:rank_tuner}.

\begin{algorithm}
\caption{A Rank Selection Algorithm (RankTuner)}
\label{alg:rank_tuner}
\begin{algorithmic}[1]
\State \textbf{Input:} Allowed rank range $[\eta_{\text{min}}, \eta_{\text{max}}]$, initial rank $\eta_0$, initial direction $d$ (\texttt{+1} or \texttt{-1}).
\State Initialize $\eta \leftarrow \eta_0$, $d \leftarrow \texttt{+1}$, previous ADU $\text{ADU}_{\text{prev}} \leftarrow 0$.
\For{each fine-tuning iteration}
    \State Fine-tune model with rank $rank$ and compute current ADU $\text{ADU}_{\text{curr}}$.
    \If{$\text{ADU}_{\text{curr}} > \text{ADU}_{\text{prev}}$}
        \State $\eta \leftarrow \eta + d$ {\Comment{Keep direction and adjust rank.}}
    \Else
        \State $d \leftarrow -d$ {\Comment{Reverse direction.}}
        \State $\eta \leftarrow \eta + d$ {\Comment{Revert to previous rank.}}
    \EndIf
    \State $\eta \leftarrow \max(\eta_{\text{min}}, \min(\eta, \eta_{\text{max}}))$ 
    \State Update $\text{ADU}_{\text{prev}} \leftarrow \text{ADU}_{\text{curr}}$.
\EndFor
\end{algorithmic}
\end{algorithm}

The RankTuner 
effectively balances fine-tuning accuracy and time efficiency by dynamically adapting the rank based on real-time feedback. This mechanism ensures that the fine-tuning process remains efficient and yields high-quality results across various tasks and environmental conditions.

\subsection{Training}

Our MARL model is based on multi-agent proximal policy optimization (MAPPO) \cite{neurips22}. 
Each agent has a flag bit to indicate whether the agent is available or not. We ignore the output of the action by non-available agents. Due to the fewer dispatching destinations in the boundary grids than in the non-boundary grids, we mask the corresponding dispatching action for agents in boundary grids. For every policy update step, we collect a batch of trajectories from the environment and compute the loss function according to \cite{PPO}. Our MARL model is trained online because online training has a higher utilization rate of sampled data. 
We apply some implementation techniques, including Generalized Advantage Estimation (GAE) with advantage normalization and value-clipping.

\section{Evaluation}
In this section, we assess the efficacy of the proposed methodology through a comprehensive evaluation. We conducted experiments based on real-world datasets of orders and compared the results with several comparison algorithms. The experimental results show that our method has better performance.

\subsection{Experiment Settings}

\noindent\textbf{Model configurations.} Our algorithm is implemented by PyTorch and DGL. The actor network of each agent is a three-layer fully connected network, where the size of the hidden layer is set to 64. The structure of the Critic network is the same as that of the actor network. The hidden layer dimension of R-GCN is 128, and the output embedding state dimension is 10. To speed up the training process, we adopted a parameter-sharing method commonly used in multi-agent training, which can reduce the computationalas overhead in experiments.

\noindent\textbf{UFM fine-tuning task configurations.} 
In the experimental settings, we consider three representative fine-tuning tasks, each designed to evaluate the effectiveness of our approach across diverse model architectures and datasets:  
\begin{itemize}
\item[$\bullet$] Image Classification Task: This task leverages the ViT pre-trained model, fine-tuned on the CIFAR-100 dataset~\cite{krizhevsky2009learning} using the LoRA technique.
\end{itemize}
\begin{itemize}
\item[$\bullet$] Image Segmentation Task: For image segmentation, we employ the SAM pre-trained model, fine-tuned on a satellite imagery dataset~\cite{ji2018fully} using LoRA.
\end{itemize}
\begin{itemize}
\item[$\bullet$] Object Detection Task: This task utilizes the YOLOv7~\cite{wang2023yolov7} model, fine-tuned on a vehicle detection dataset~\cite{vehicle-detection-nmzlp_dataset}. The dataset contains annotated images of vehicles in various environments, emphasizing diverse perspectives, sizes, and contexts. As YOLO fine-tuning involves full fine-tuning, rank selection is not applicable, and RankTuner is not utilized for this task.
\end{itemize}
Each fine-tuning configuration was carefully crafted to reflect real-world scenarios, emphasizing the utility of our framework in handling diverse data modalities and model architectures while maintaining computational efficiency.


\noindent\textbf{Order serving dataset.}
The detailed information of the ride order settings comes from the New York City Taxi dataset \cite{newyork_taxi}. The real-world dataset records the travel data of taxis in New York, including the latitude and longitude of passengers boarding and alighting, the start and end time of each order, the fare paid by passengers, and so on. 

\begin{figure}[t]
\centering  
\subfigure[Distribution of orders.]{
\label{fig: sub.1 order distribution}
\includegraphics[width=0.23\textwidth]{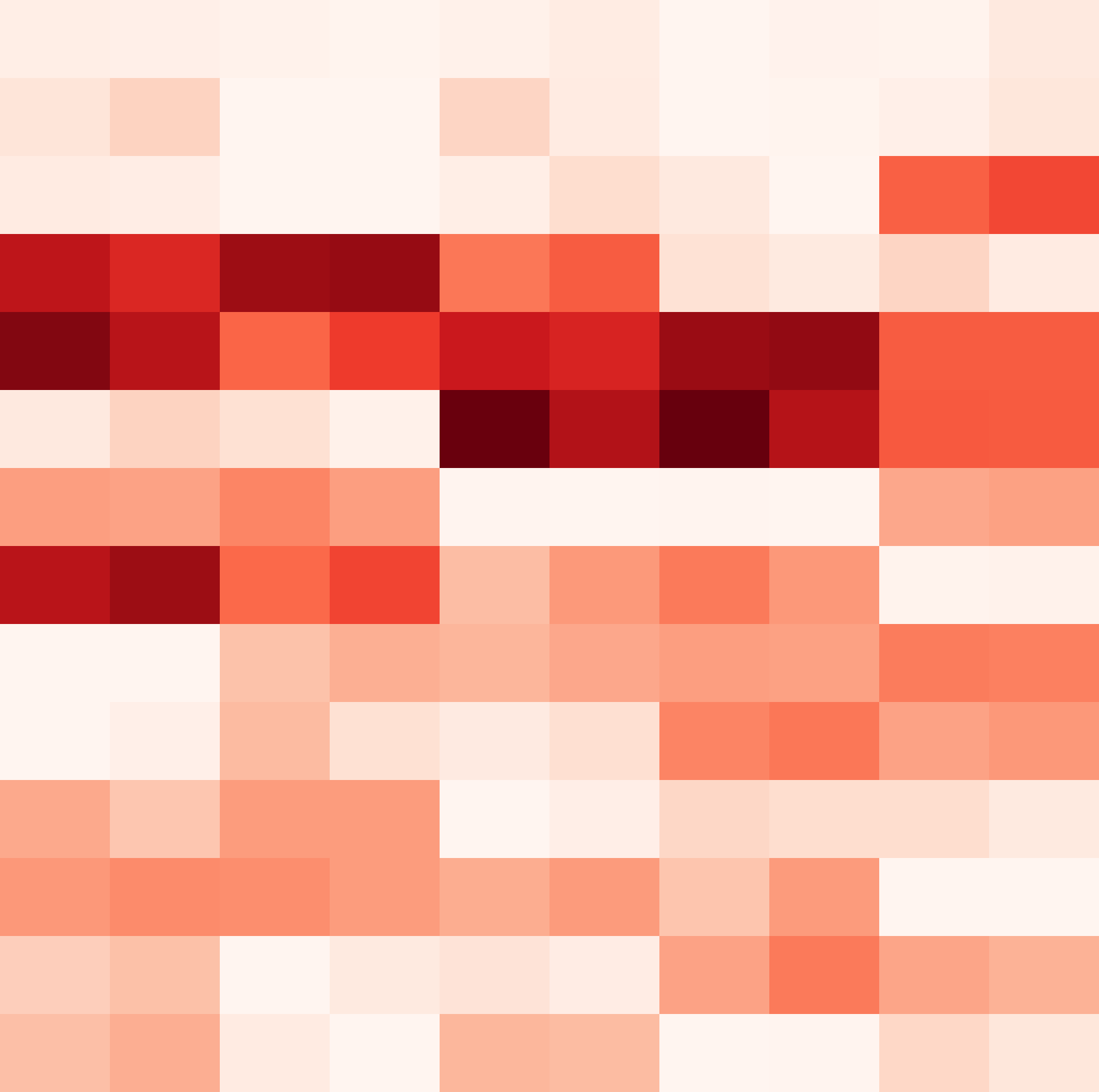} }
\hfil
\subfigure[Distribution 1 of PoIs]{
\label{fig: sub.2 poi distribution 1}
\includegraphics[width=0.23\textwidth]{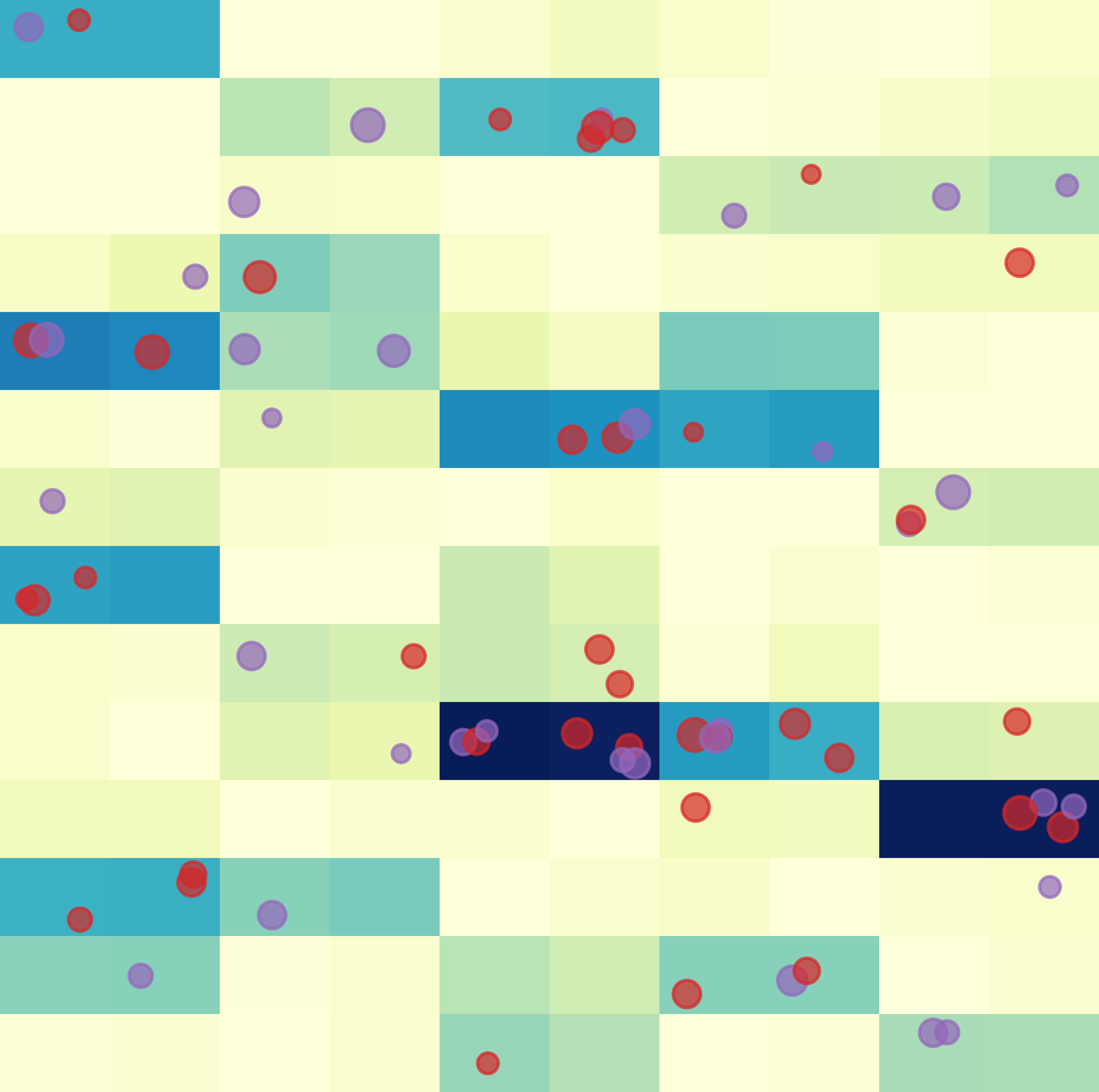}} 

\subfigure[Distribution 2 of PoIs]{
\label{fig: sub.2 poi distribution 2}
\includegraphics[width=0.23\textwidth]{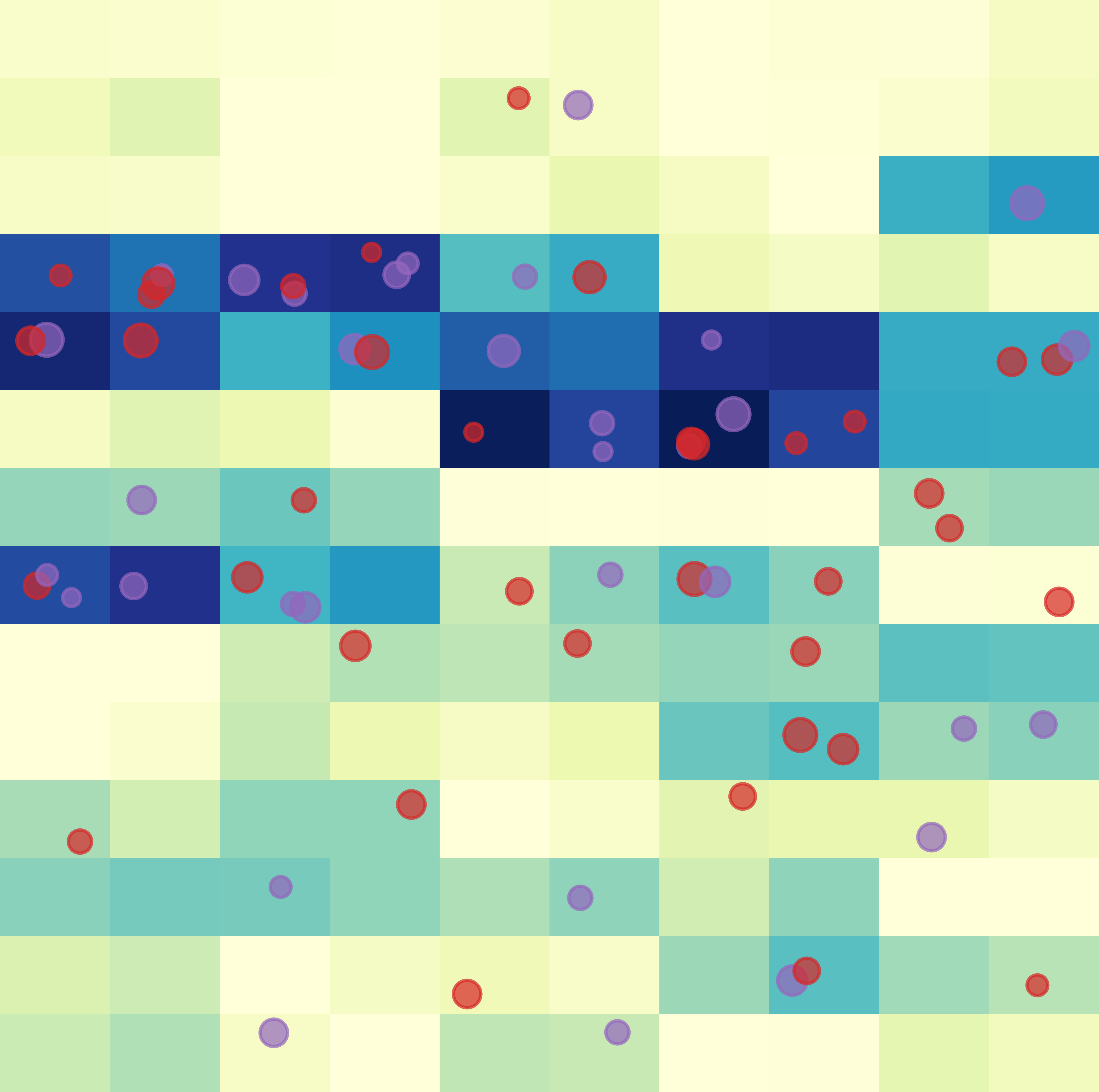}}
\hfil
\subfigure[Distribution 3 of PoIs]{
\label{fig: sub.2 poi distribution 3}
\includegraphics[width=0.23\textwidth]{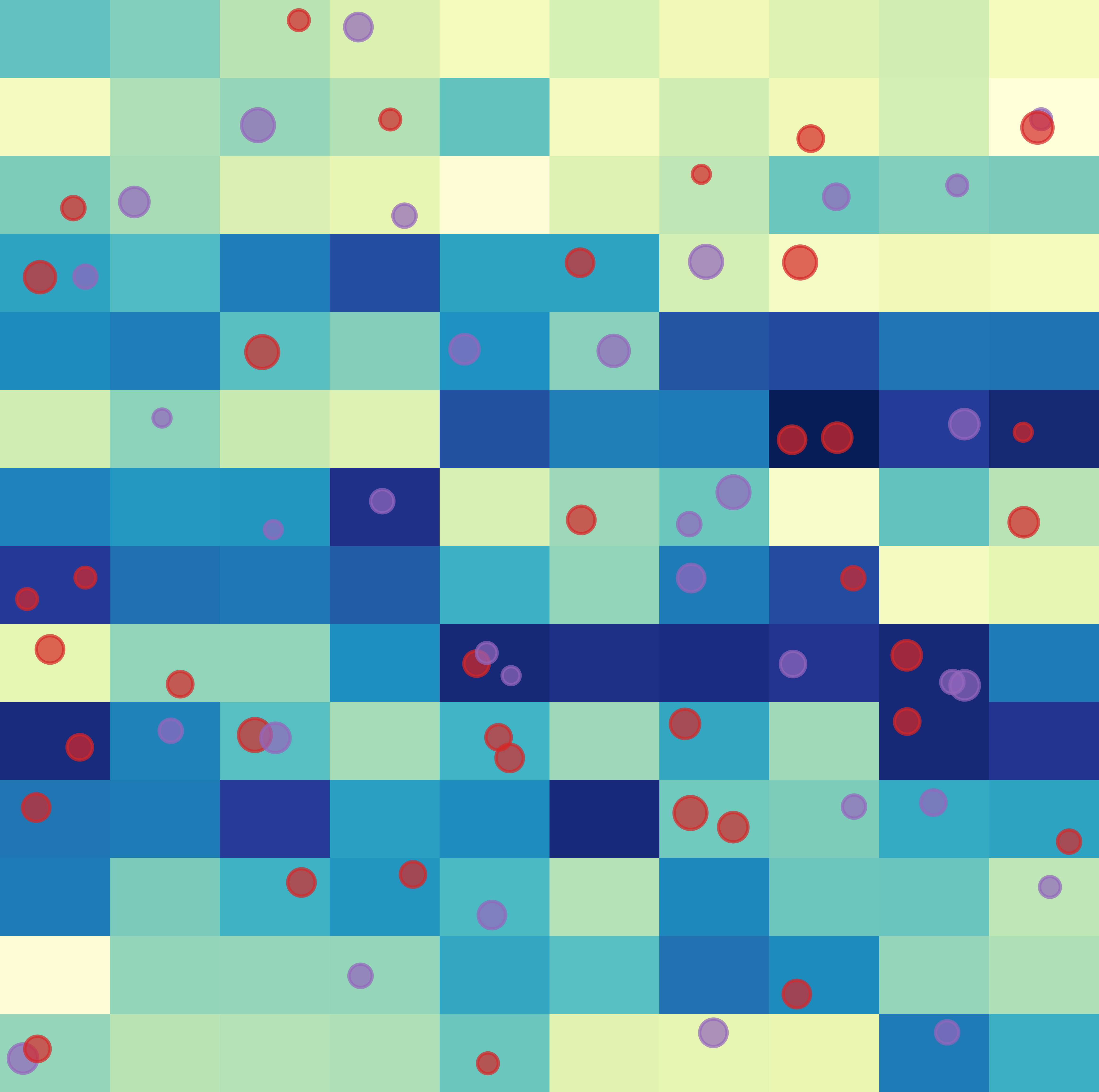}} 

\caption{
The distribution probability of orders and PoIs in the target area. Darker grid colors indicate higher probabilities of generating orders or PoIs. PoIs are shown as points, with colors representing task types: red for image classification and purple for image segmentation. Point sizes are proportional to data volumes, reflecting task intensity.}
\label{fig: dis of order and poi}
\end{figure}

\noindent\textbf{Simulator design.}
At the beginning of each time slot, the simulator dynamically generates ride requests and PoIs. The data volume associated with each PoI is generated by random sampling within a range of 3 to 12 packages, while we manually construct a probability distribution for the PoIs. We have defined three distinct probability distributions for PoIs, labeled as distribution 1, distribution 2, and distribution 3. The distribution of orders and PoI is shown in Fig. \ref{fig: dis of order and poi}. Distribution 1 exhibits a significant disparity between the distributions of PoIs and ride orders. As for distribution 2, PoIs follow a distribution similar to that of orders. Distribution 3 indicates a uniform distribution of PoIs within the target area. Our fleet includes 100 ride-hailing vehicles, each with a set data collection rate of 1 package per time slot. To simplify the problem, we assume that the vehicle can reach the dispatch destination after one time slot like most research does \cite{KDD18, assume_cost_one_time_slot_to_reach_dispatched_destination}. We set that if an order is not accepted by a vehicle within 15 time slots from the time it is generated, the order will be invalid. 

\begin{figure}[t]
\centerline{\includegraphics[width=1.0\linewidth]{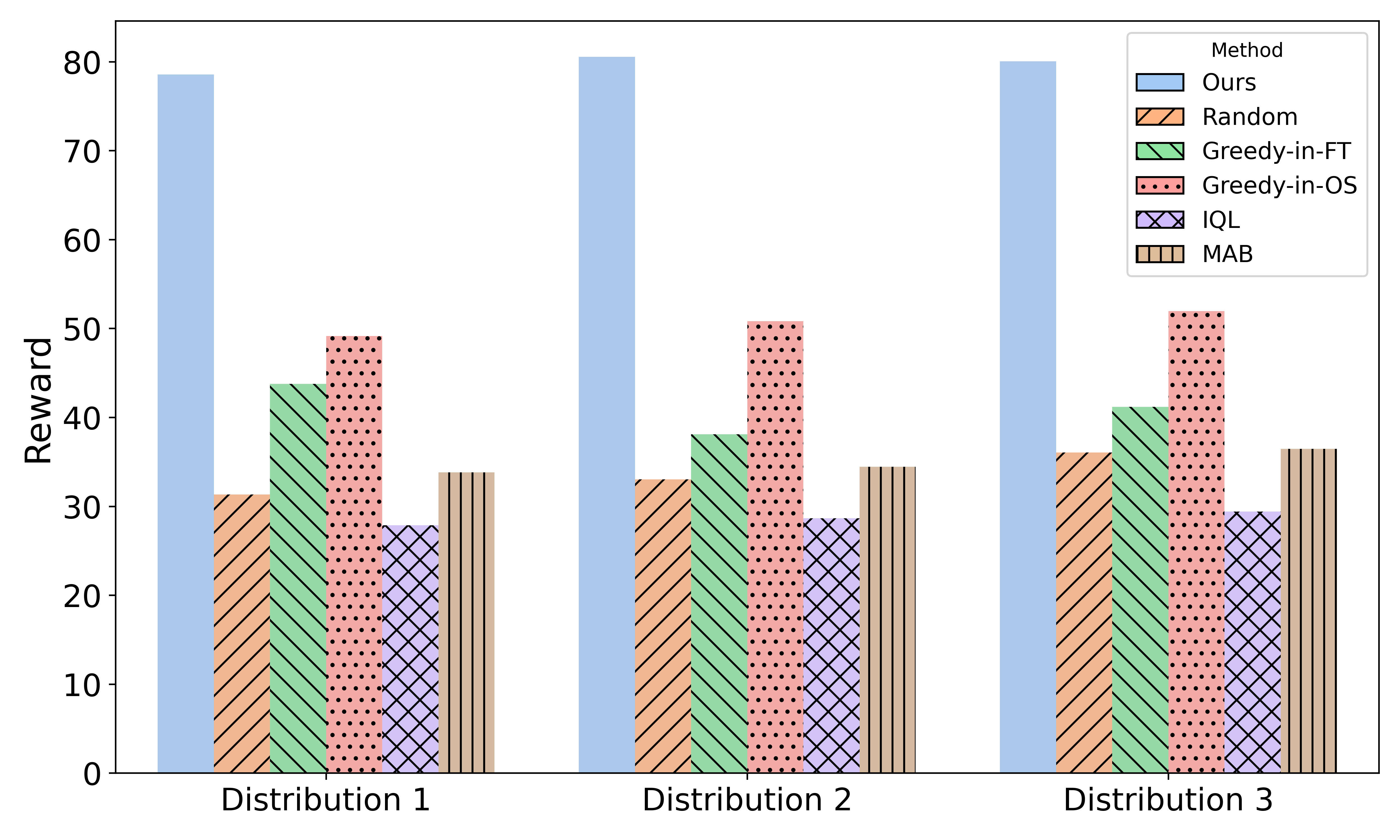}}
\caption{Different results of different distributions. }
\label{different result of different distribution}
\end{figure}

\noindent\textbf{Baselines.}
We consider the following baselines for our algorithm to compare.
\begin{itemize}
\item[$\bullet$] 
Random: The vehicle randomly selects actions from a set of actionable actions based on the grid it is currently in.
\end{itemize}

\begin{itemize}
\item[$\bullet$] 
Greedy-in-OS (with OS representing order serving): If there are orders distributed in the grid, available vehicles will choose to prioritize accepting orders. If there are no orders available, vehicles will go to other grids, collect data or stay.
\end{itemize}

\begin{itemize}
\item[$\bullet$] 
Greedy-in-FT (with FT representing fine-tuning): When fine-tuning tasks are associated with PoIs within a grid, available vehicles prioritize data collection from these PoIs. In the absence of such PoIs, vehicles adapt by either traveling to adjacent grids, fulfilling passenger orders, or remaining stationary.
\end{itemize}

\begin{itemize}
\item[$\bullet$] 
Multi-armed bandits (MAB): For any action vector, MAB uses the sum of the empirical average reward and an upper-confidence bound as the final reward function \cite{UCB}. 
\end{itemize}

\begin{itemize}
\item[$\bullet$] 
IQL: IQL extends DQN to a decentralized multi-agent reinforcement learning environment, with the ability to handle high-dimensional and complex environments.
\end{itemize}


Our performance evaluation metrics include 1) QoS, which reflects the overall performance of our framework. 2) ADI, which reflects the quality of order-serving, 3) ADU, which reflects the quality of data collection and 4) the average AoI, which is the average of the AoI of the collected PoIs and reflects the overall data freshness.

\subsection{Performance}
We analyze the experimental results as follows. The distribution of PoI is distribution 1.

\begin{figure*}[htbp] 
	\centering  
	\vspace{-0.35cm} 
	\subfigtopskip=2pt 
	\subfigbottomskip=2pt 
	\subfigcapskip=-5pt 
	\subfigure[QoS comparison with baselines.]{
		\label{QoS}
		\includegraphics[width=0.32\linewidth]{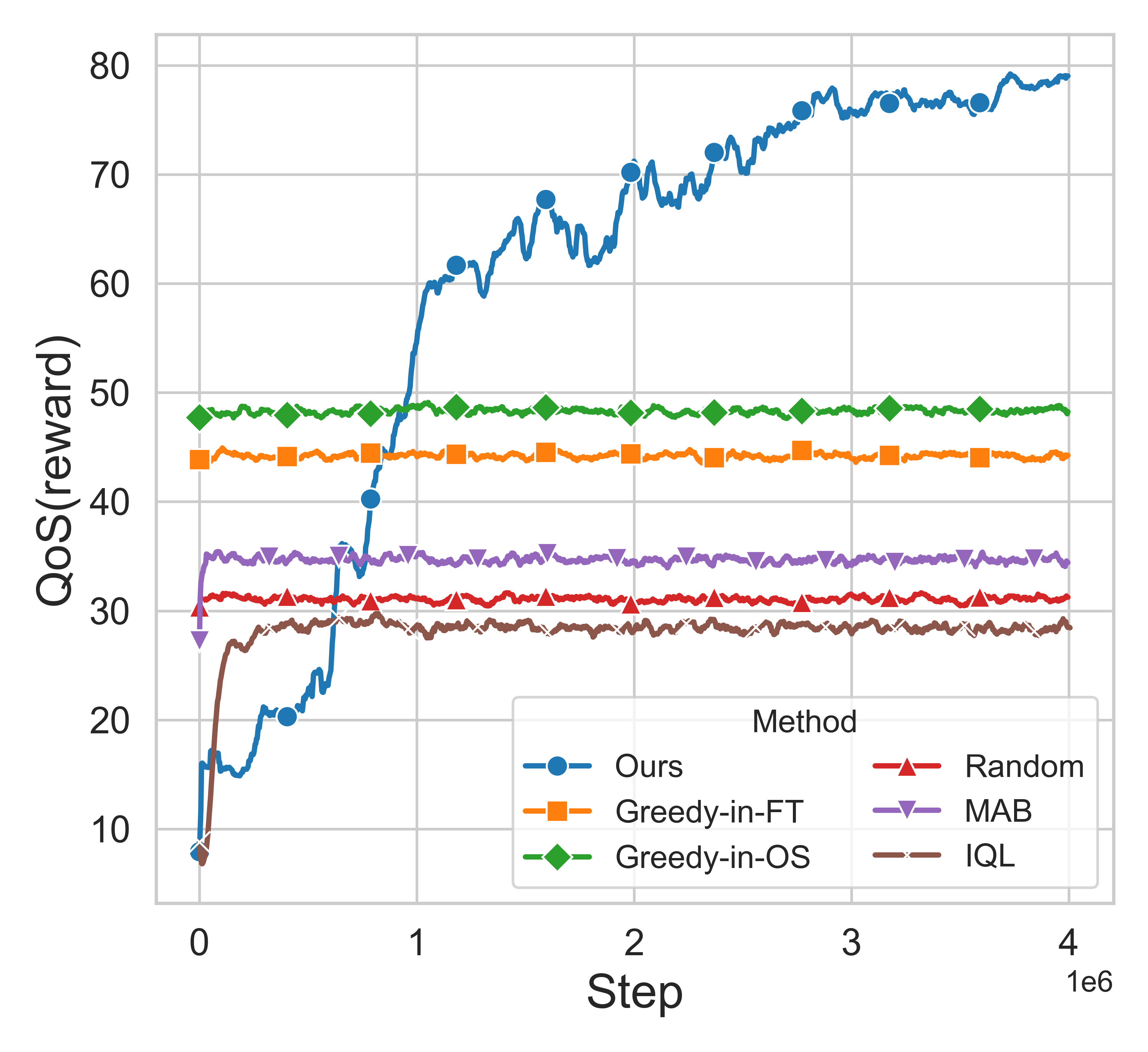}}
	\subfigure[ADI comparison with baselines.]{
		\label{ADI}
		\includegraphics[width=0.32\linewidth]{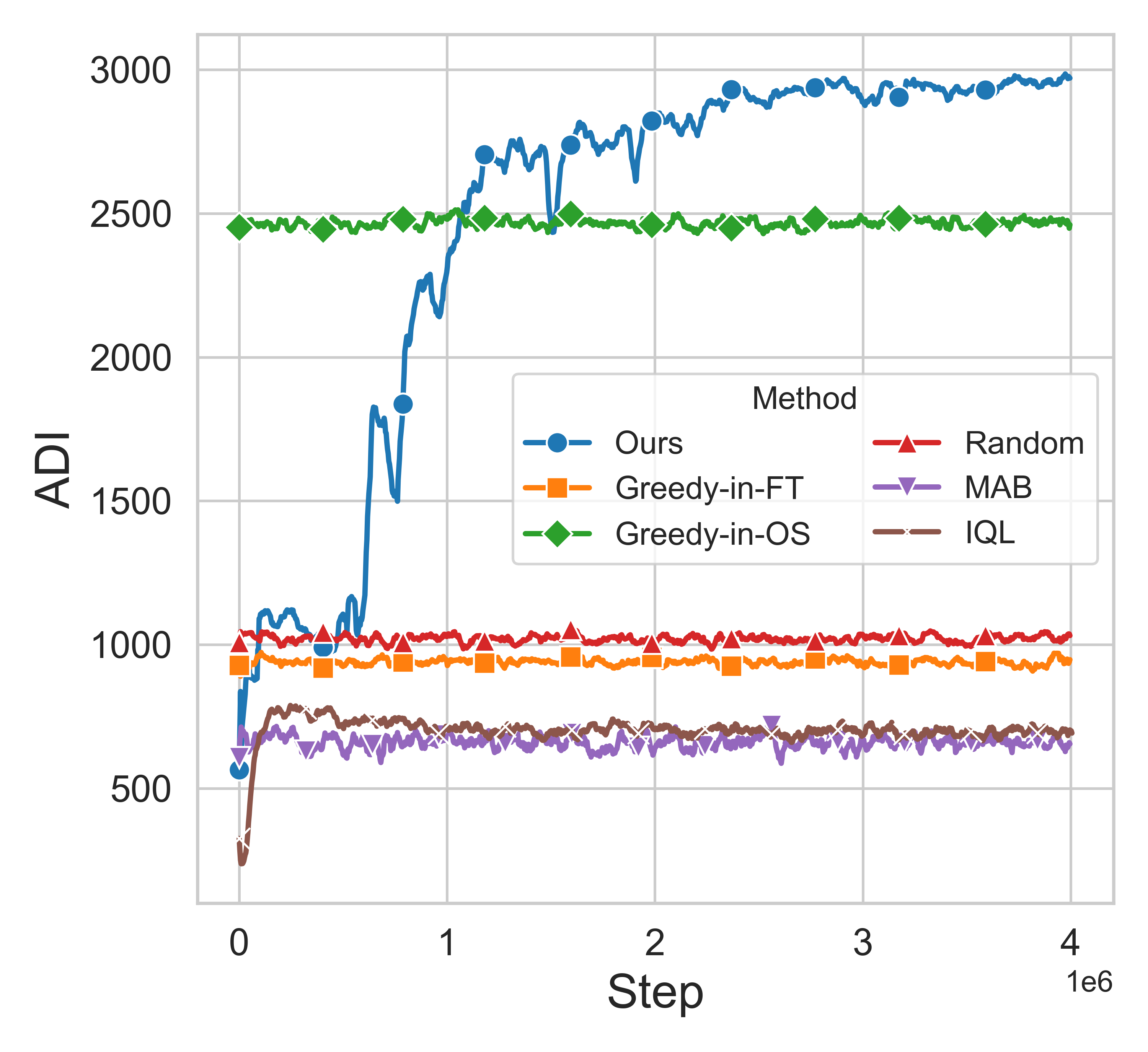}}
        \subfigure[ADU comparison with baselines.]{
		\label{ADU}
		\includegraphics[width=0.32\linewidth]{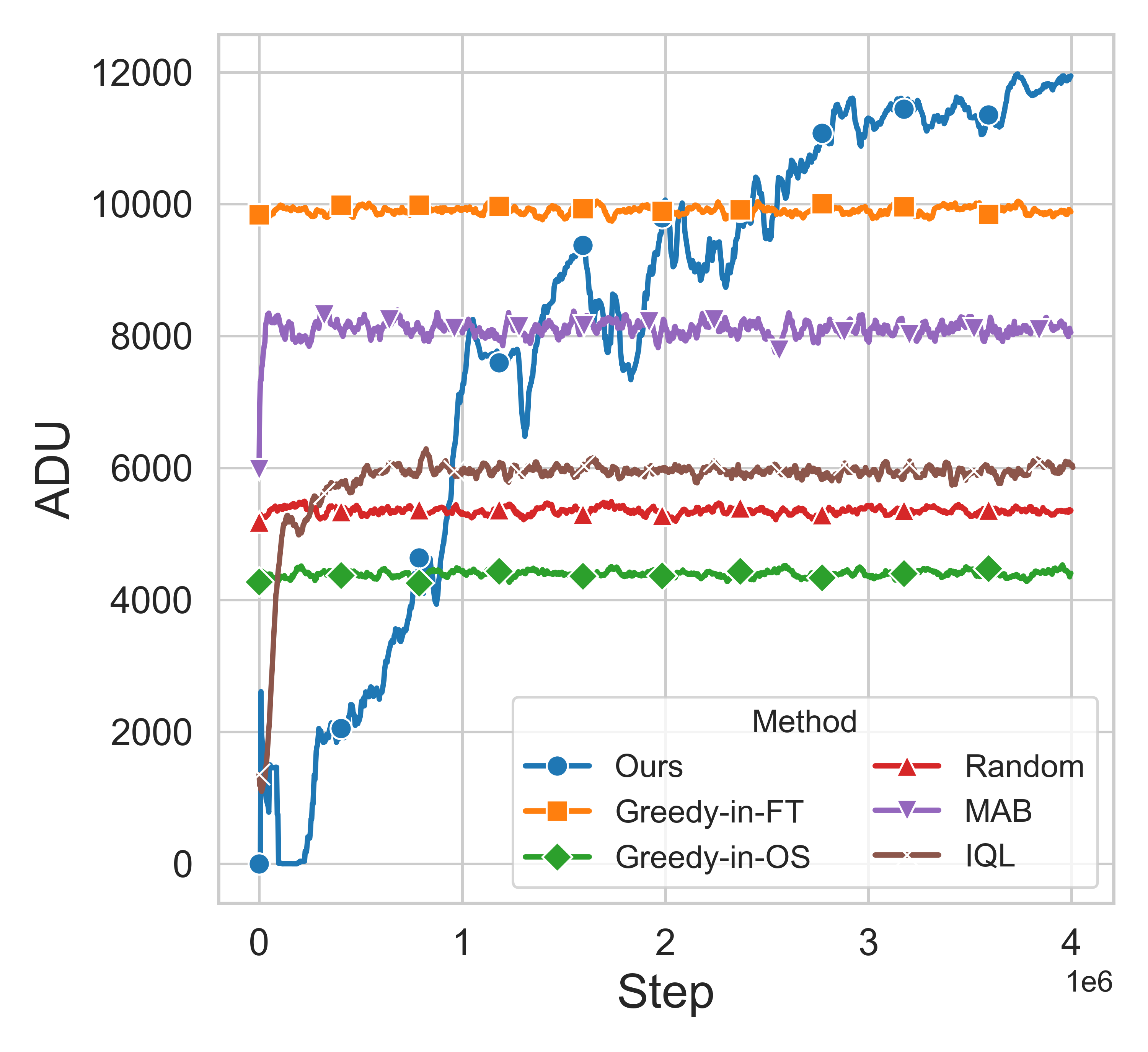}}
	\caption{Performance comparison with baselines using Distribution 1 of the data}
	\label{contrast}
\end{figure*}

1) QoS: We demonstrate the QoS of our method and the other 5 baselines during the training in Fig. \ref{QoS}. The experimental results indicate that our framework outperforms all baselines in terms of QoS. Although we lead to GNN, which increases the complexity of the model to some extent, our algorithm still maintains satisfactory convergence. MAB does not consider the state of the environment and the results of each interaction are independent of past actions, causing poor QoS performance in complex environments. Under the IQL algorithm, agents that make independent decisions based on local states are difficult to adapt to environments with high randomness and complexity, resulting in their inability to learn better decisions.

\begin{figure}[t]
\centerline{\includegraphics[width=0.7\linewidth]{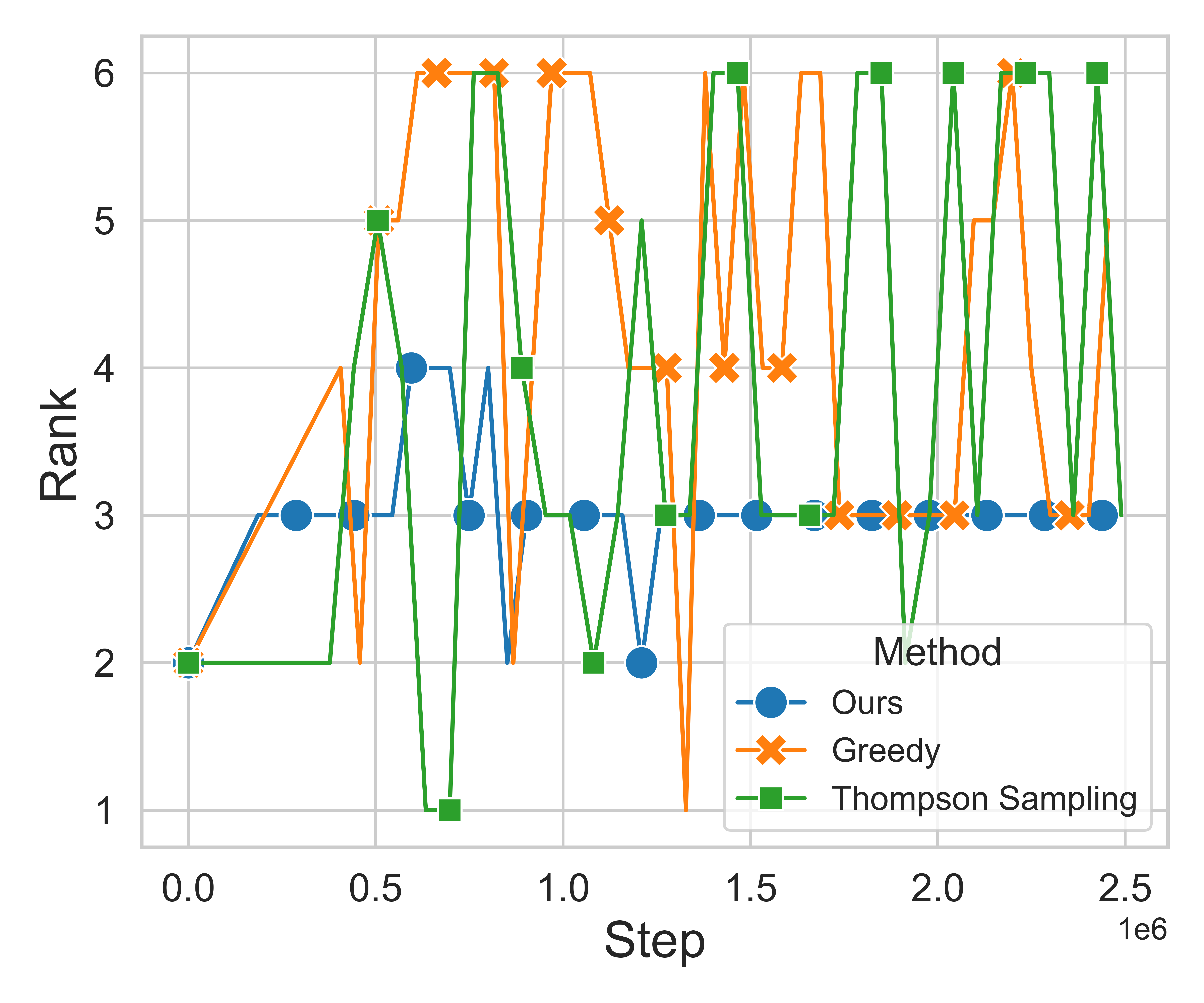}}
\caption{Rank Variation Over Training Steps.
}
\label{fig:rank}
\end{figure}

\begin{table}[ht]
\centering
\caption{QoS Comparison of Different Rank Strategies}
\label{tab:qos_comparison}
\resizebox{0.61\linewidth}{!}{ 
\begin{tabular}{|c|c|}
\hline
\textbf{Method} & \textbf{QoS} \\ \hline
Ours            & 72.78        \\ \hline
Greedy          & 68.93        \\ \hline
Thompson Sampling & 63.49      \\ \hline
\end{tabular}
}
\end{table}

2) ADI: 
We recorded the ADI during the training process and displayed it in Fig. \ref{ADI}. As the number of training steps increases, the ADI obtained based on our algorithm continues to rise and eventually converges around 2.5 million steps. In contrast to the baseline algorithms, our algorithm achieves the maximum income. The ADI obtained by Greedy-in-OS is higher than that based on Greedy-in-FT, which is in line with our expectations because if the priority of accepting orders is higher, the frequency of vehicles choosing to accept orders will be higher compared to collecting data. Our algorithm not only learns the distribution of orders and schedules vehicles based on it but also selects the best time to accept orders, which makes our method superior to these baselines in order income.
\begin{table}[ht]
\centering
\caption{The average accuracy of different methods.
}
\resizebox{0.61\linewidth}{!}{
\begin{tabular}{|c|c|}
\hline
\textbf{Method} & \textbf{Accuracy(\%)} \\
\hline
Ours   & 69.03 \\
Random & 60.77 \\
Greedy-in-FT    & 62.97 \\
Greedy-in-OS  & 60.14 \\
IQL    & 66.48 \\
MAB    & 65.59 \\
\hline
\end{tabular}
}

\label{tab:accuracy_methods}
\end{table}
\begin{table*}[ht]
\centering
\caption{Performance on Multiple Tasks (Image Classification, Object Detection, Image Segmentation)}
\resizebox{\linewidth}{!}{
\begin{tabular}{|c|c|c|c|c|c|c|c|c|c|c|c|}
\hline
\textbf{Method} & \multicolumn{3}{c|}{\textbf{One Task}} & \multicolumn{3}{c|}{\textbf{Two Tasks}} & \multicolumn{3}{c|}{\textbf{Three Tasks}} \\
\cline{2-10}
                & \textbf{QoS} & \textbf{ADI} & \textbf{ADU} & \textbf{QoS} & \textbf{ADI} & \textbf{ADU} & \textbf{QoS} & \textbf{ADI} & \textbf{ADU} \\
\hline
\textbf{Ours}   & \textbf{66.24} & \textbf{2,834.13} & \textbf{8,474.18} & \textbf{78.57} & \textbf{2,943.96} & \textbf{11,914.06} & \textbf{67.97} & \textbf{2,896.53} & \textbf{8,747.41} \\
Random          & 30.07         & 1,009.86         & 5,081.45          & 31.35          & 1,014.28         & 5,469.35          & 30.75          & 1,009.86         & 5,300.96          \\
greedy-in-FT    & 36.92          & 965.66          & 7,465.04          & 43.78          & 993.40          & 9,527.69          & 37.74          & 1,000.57          & 7,570.97          \\
greedy-in-OS    & 47.49          & 2,437.77         & 4,262.12          & 49.16          & 2,540.21         & 4,338.30          & 48.02          & 2,437.77         & 4,432.99          \\
IQL             & 26.57          & 733.97          & 5,199.41          & 27.89          & 650.55          & 5,680.78          & 26.58          & 649.82          & 5,578.69          \\
MAB             & 34.19          & 784.63          & 7,402.70          & 33.83          & 571.07          & 8,241.65          & 34.54          & 683.61          & 7,965.09          \\
\hline
\end{tabular}
}
\label{tab:task scalability}
\end{table*}
3) ADU: According to Fig. \ref{ADU}, our algorithm also achieved the best performance in data collection. Although the MAB algorithm also has good performance in ADU, combined with Fig. \ref{ADI} and Fig. \ref{ADU}, it doesn't learn the effects of accepting orders and is overly focused on data collection. As the training progresses, the ADU of IQL continuously increases and converges to a local optimum.

4) 
The average accuracy:
To assess the impact of data freshness and quantity on model fine-tuning, we evaluate the inference accuracy achieved on the test set after fine-tuning, as shown in Table \ref{tab:accuracy_methods}. Our approach achieves the highest accuracy, highlighting the advantages of effectively managing both data freshness and volume during the fine-tuning process. In comparison, baseline methods such as Random, Greedy-in-FT, and Greedy-in-OS strategies exhibit lower accuracy due to their suboptimal handling of data freshness and quantity. While the Greedy-in-FT strategy initially reduces AoI, its inability to implement an effective dispatching strategy limits its ability to optimize data collection and fine-tuning accuracy. Our method, which incorporates GNNs to fuse PoI attributes, captures a more detailed and dynamic state representation, enabling better utilization of data freshness and quantity, and thereby outperforming methods like IQL, which lacks sufficient PoI information and struggles to fully capture the sensitivity of data value to freshness. 


Overall, our algorithm achieves a balance between order income and data utility, ensuring the completion quality of both tasks as much as possible through reasonable decision-making in the context of limited vehicle numbers.

\subsection{Effect of RankTuner}
Our experiments evaluate the effectiveness of the RankTuner by comparing it with two alternative rank selection strategies:
\begin{itemize}
\item[$\bullet$] 
Greedy: It adjusts the LoRA rank by comparing the ADU of the current step with the previous step. If the ADU improves, the rank is further adjusted in the same direction; otherwise, no change is made.
\end{itemize}
\begin{itemize}
\item[$\bullet$] 
Thompson sampling~\cite{chapelle2011empirical}: This strategy employs a probabilistic method to balance exploration and exploitation. By sampling ranks according to their estimated likelihood of enhancing ADU, it facilitates broader exploration of potential rank configurations.
\end{itemize}
According to Table~\ref{tab:rank}, rank 3 provides the best trade-off between fine-tuning time and accuracy. While rank 6 achieves the highest precision, its fine-tuning time is nearly double that of rank 3, limiting the number of fine-tuning tasks that can be completed within a given timeframe. The reduced fine-tuning time of rank 3 enables agents to perform more tasks, improving overall ADU and reinforcing the need for a strategy that balances efficiency and accuracy. Fig.~\ref{fig:rank} demonstrates that our RankTuner converges to the optimal rank 3 faster than the Greedy and Thompson Sampling strategies. The Greedy strategy often suffers from limited exploration, resulting in slower adaptation to dynamic environments. Thompson Sampling, while more exploratory, is prone to settling into local optima when the benefits of the local and global solutions are close~\cite{phan2019thompson}. This issue arises because Thompson Sampling assigns higher probabilities to seemingly optimal ranks early on, making it challenging to escape suboptimal configurations. As shown in Table~\ref{tab:qos_comparison}, the RankTuner achieves the highest QoS after 2.5 million training steps, outperforming both baselines. These results validate the RankTuner's ability to dynamically adjust ranks, balance exploration and exploitation, and optimize fine-tuning tasks in diverse and uncertain environments.

\subsection{Impact of varying distribution of PoIs} 
To verify the robustness of the algorithm under different distributions, we set different distribution types for PoI in our experiment. As shown in Fig. \ref{different result of different distribution}, our method exhibits better performance compared to other baselines under different PoI distributions. We find that when the distribution of PoIs is roughly consistent with the distribution of orders, the total QoS obtained by the fleet will be higher. We speculate that this is because the consistent distribution reduces the complexity of the algorithm in learning the distribution of the environment and making dispatching decisions. There are some grids within which there are a considerable number of orders and PoI simultaneously, allowing for better maximization of total QoS by scheduling vehicles to these types of grids. Compared to other distribution types, the rule-based methods of random, greedy on orders, and greedy on PoIs achieve higher QoS when PoIs are uniformly distributed. This is mainly because in the case of uniformly distributed PoIs, there is a similar amount of PoIs waiting to be collected in each grid. These rule-based methods generate dispatching actions randomly, making the probability of vehicles traveling to each grid roughly the same, thereby increasing the probability of vehicles collecting PoI data from each grid.

\subsection{Scalability: impact of the number of tasks}
In this section, we analyze the scalability of our proposed method in terms of the number of tasks, examining its performance across one, two, and three tasks. As shown in Table \ref{tab:task scalability}, we observe that as the number of tasks increases, there are variations in the metrics of QoS, ADI and ADU. Specifically, while the QoS improves when moving from one to two tasks, a slight decrease occurs when transitioning to three tasks. This decline can be attributed to the increased complexity of handling multiple task types, each with varying fine-tuning accuracy. The fine-tuning accuracy for each task can differ significantly, and some tasks may have lower maximum accuracy, leading to a reduced overall QoS. Additionally, the growing diversity of tasks introduces more challenging conditions for optimizing data collection and model fine-tuning. Despite these challenges, our method still outperforms other approaches, demonstrating its ability to efficiently manage multiple tasks while maintaining high ADI and ADU values. The stable performance across tasks highlights the scalability of our framework in environments with varying task complexities.

\section{CONCLUSION}
This paper presents a novel algorithmic framework that integrates dispatching and data collection for fine-tuning in ride-hailing vehicles. We recognize that data freshness and volume are crucial factors influencing fine-tuning accuracy. Specifically, the amount of collected data and its freshness significantly affect the accuracy of model fine-tuning, which in turn impacts the overall system performance. We model the tasks of dispatching, order-serving, and data collection for fine-tuning as a joint optimization problem, aiming to maximize the QoS. To solve this, we propose a multi-agent reinforcement learning algorithm based on the R-GCN model, enabling effective fusion of diverse environmental states and improving the efficiency of fine-tuning tasks. Our experiments demonstrate that the proposed approach effectively optimizes both ride-hailing services and data collection for fine-tuning, achieving superior results in terms of accuracy and overall system performance.


\section*{Acknowledgment}

The authors would like to thank anonymous reviewers for their insightful comments and suggestions. This work was supported by the National Science Foundation of Chinagrant(No.62102460 and No.U20A20159), Guangzhou Science and Technology Plan Project (No. 202201011392), Guangdong Basic and Applied Basic Research Foundation (No. 2023A1515012982), Young Outstanding Award under the Zhujiang Talent Plan of Guangdong Province, Guangdong Basic and Applied Basic Research Foundation (No. 2023B1515120058), and Guangzhou Basic and Applied Basic Research Program (No. 2024A04J6367). Xiaoxi Zhang is the corresponding author.



\ifCLASSOPTIONcaptionsoff
  \newpage
\fi

\bibliographystyle{IEEEtran}
\bibliography{ref}
\newpage
%
\begin{IEEEbiography}
[{\includegraphics[width=1in,height=1.25in,clip,keepaspectratio]{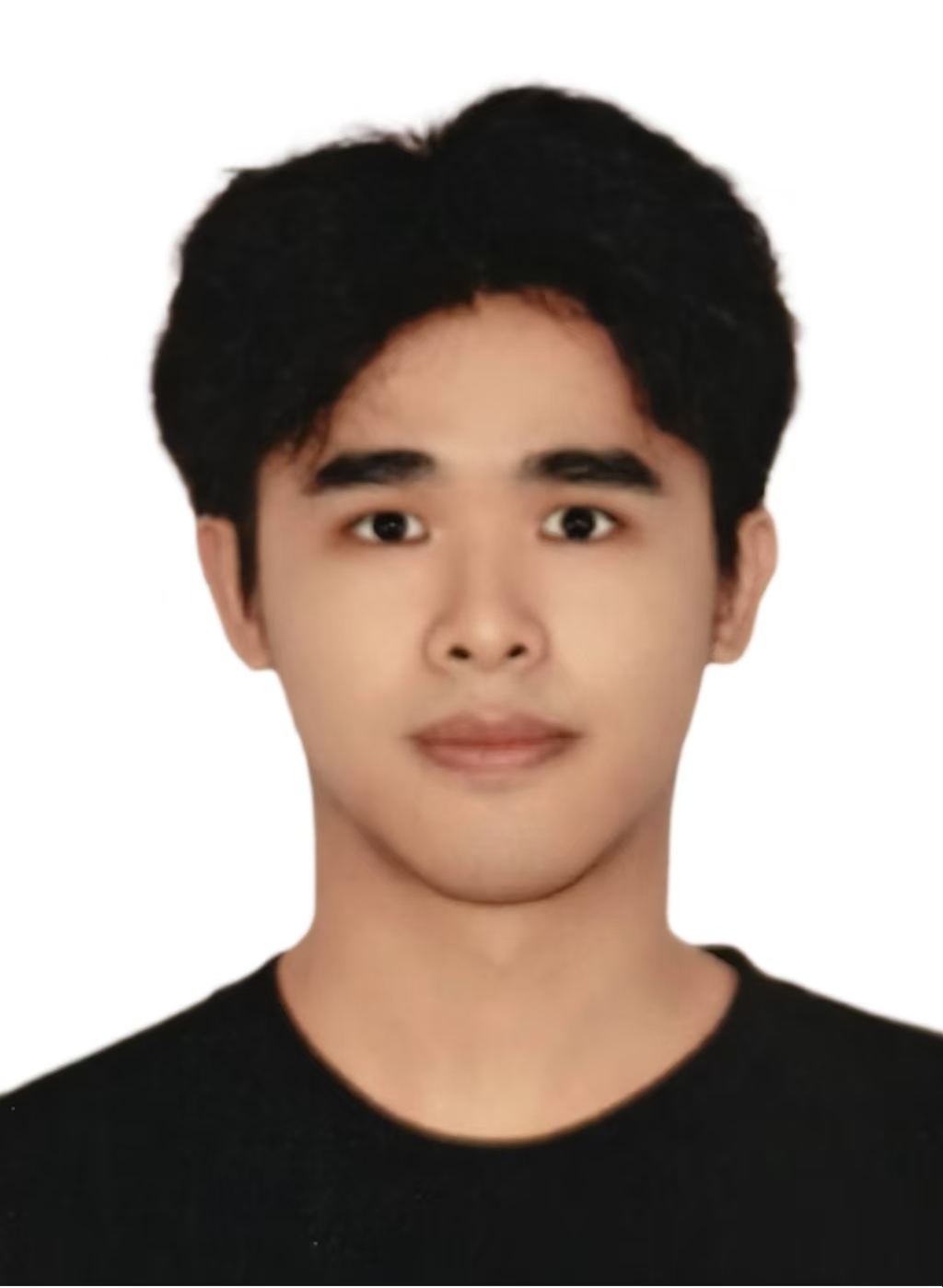}}]{Bokeng Zheng}
is currently pursuing a bachelor’s degree with the School of Computer Science and Engineering, Sun Yat-sen University. His research interests include edge computing and the application of fine-tuning.
\end{IEEEbiography}
\vskip -2\baselineskip plus -1fil
\begin{IEEEbiography}
[{\includegraphics[width=1in,height=1.25in,clip,keepaspectratio]{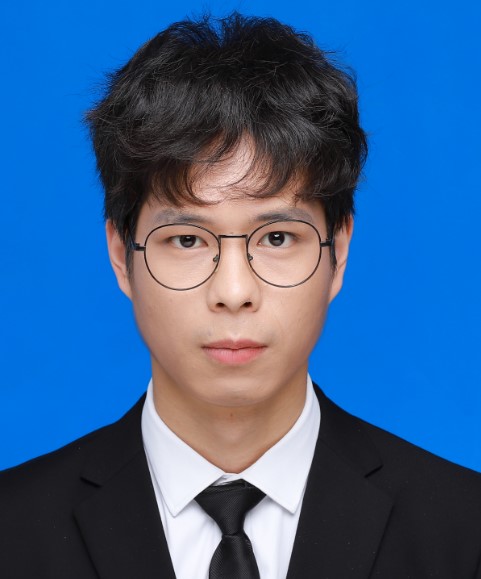}}]{Bo Rao}
received his bachelor’s degree from the School of Electronics and Information, South China University of Technology in 2022. He is currently pursuing a master’s degree with the School of Computer Science and Engineering, Sun Yat-sen University. His research interests include edge computing and the application of reinforcement learning.
\end{IEEEbiography}
\vskip -2\baselineskip plus -1fil
\begin{IEEEbiography}
[{\includegraphics[width=1in,height=1.25in,clip,keepaspectratio]{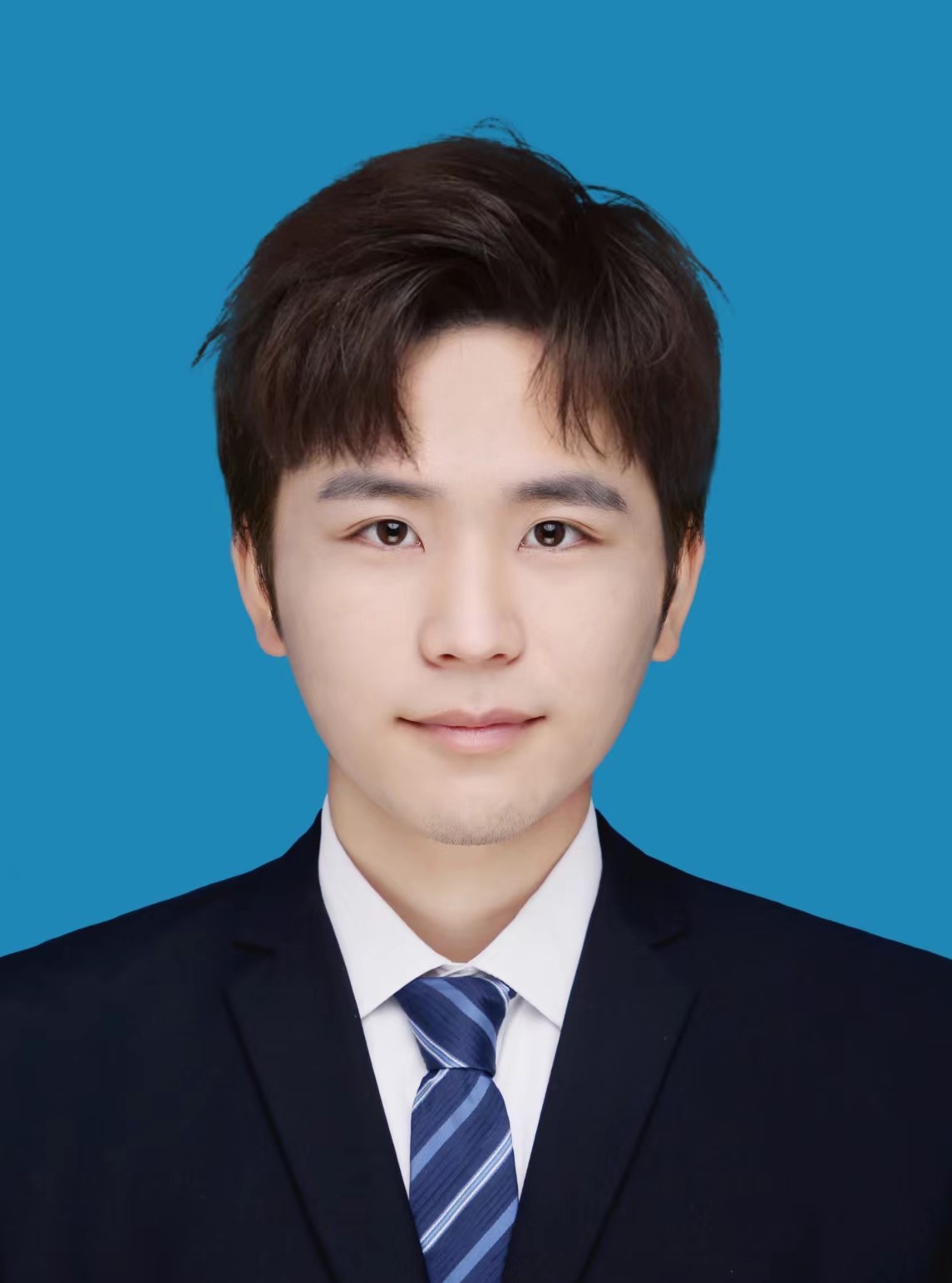}}]{Tianxiang Zhu}
received his bachelor’s degree from the College of Software Engineering, Sichuan University in 2021. He is currently pursuing a master’s degree with the School of Computer Science and Engineering, Sun Yat-sen University. His research interests include reinforcement learning and intelligent manufacturing.
\end{IEEEbiography}


\vskip -2\baselineskip plus -1fil
\begin{IEEEbiography}[{\includegraphics[width=1in,height=1.25in,clip,keepaspectratio]{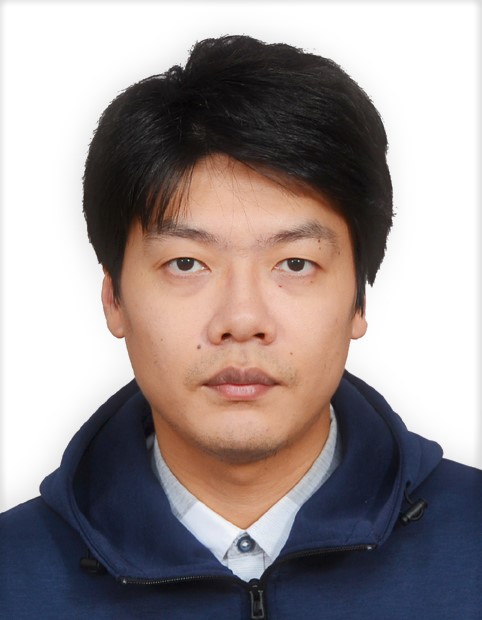}}]{Jingpu Duan}
 received the B.E. degree from the Huazhong University of Science and Technology, Wuhan, China, in 2013, and the Ph.D. degree from the University of Hong Kong, Hong Kong, China, in 2018. He is currently a Research Assistant Professor with the Institute of Future Networks, Southern University of Science and Technology, Shenzhen, China. He also works with the Department of Communications, Pengcheng Laboratory, Shenzhen, China. His research interest includes designing and implementing high-performance networking systems.
\end{IEEEbiography}
\vskip -2\baselineskip plus -1fil

\begin{IEEEbiography}[{\includegraphics[width=1in,height=1.25in,clip,keepaspectratio]{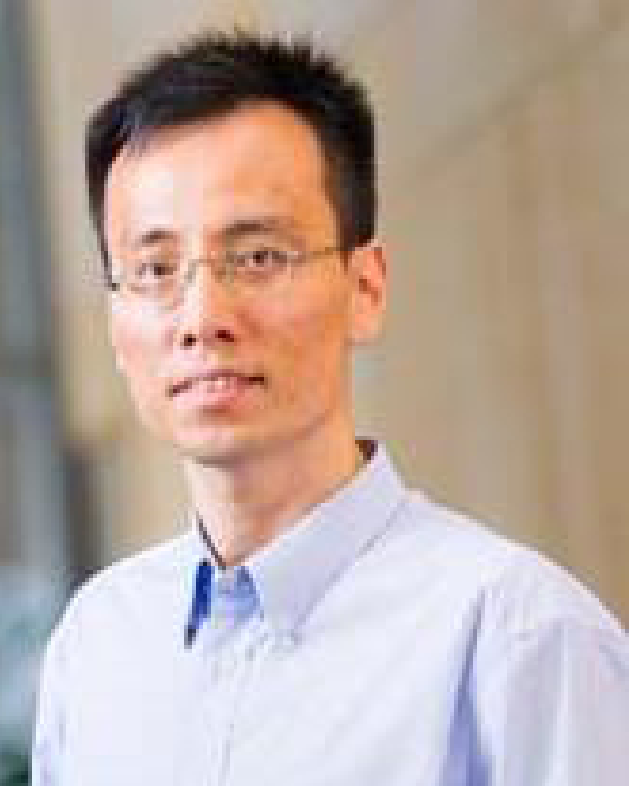}}]{Chee Wei Tan}
received an M.A. and Ph.D. in
Electrical Engineering from Princeton University. He
is an Associate Professor of Computer Science and
Engineering at Nanyang Technological University.
He conducts research in networks, distributed optimization, and generative AI. Dr. Tan has served as IEEE Distinguished Lecturer and Editor for IEEE Transactions on Cognitive Communications and Networking, IEEE/ACM Transactions on Networking,
and IEEE Transactions on Communications. He has
received the Princeton University Wu Prize for Excellence, the Google Faculty Award, and several teaching excellence awards.
He was selected twice for the U.S. National Academy of Engineering ChinaAmerica Frontiers of Engineering Symposium. He is a Co-Chair of the
Cognitive Radio and AI-Enabled Networks Symposium at IEEE GLOBECOM 2025 and a member of ACM Learning at Scale Extended Steering Committee.
\end{IEEEbiography}
\vskip -2\baselineskip plus -1fil
\begin{IEEEbiography}[{\includegraphics[width=1in,height=1.25in,clip,keepaspectratio]{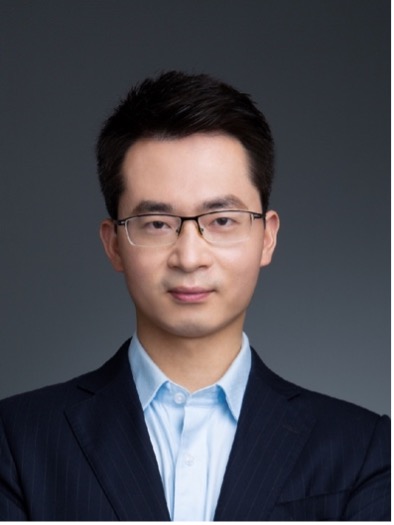}}]{Zhi Zhou} 
 received the B.S., M.E., and Ph.D. degrees in 2012, 2014, and 2017, respectively, all from the School of Computer Science and Technology at Huazhong University of Science and Technology (HUST), Wuhan, China. He is currently an associate professor in the School of Data and Computer Science at Sun Yat-sen University, Guangzhou, China. In 2016, he was a visiting scholar at University of Gottingen. He was nominated for the ¨2019 China Computer Federation CCF Outstanding Doctoral Dissertation Award, the sole recipient of the 2018 ACM Wuhan Hubei Computer Society Doctoral Dissertation Award, and a recipient of the Best Paper Award of IEEE UIC 2018. His research interests include edge computing, cloud computing, and distributed systems.
\end{IEEEbiography}
 
\vskip -2\baselineskip plus -1fil
\begin{IEEEbiography}[{\includegraphics[width=1in,height=1.25in,clip,keepaspectratio]{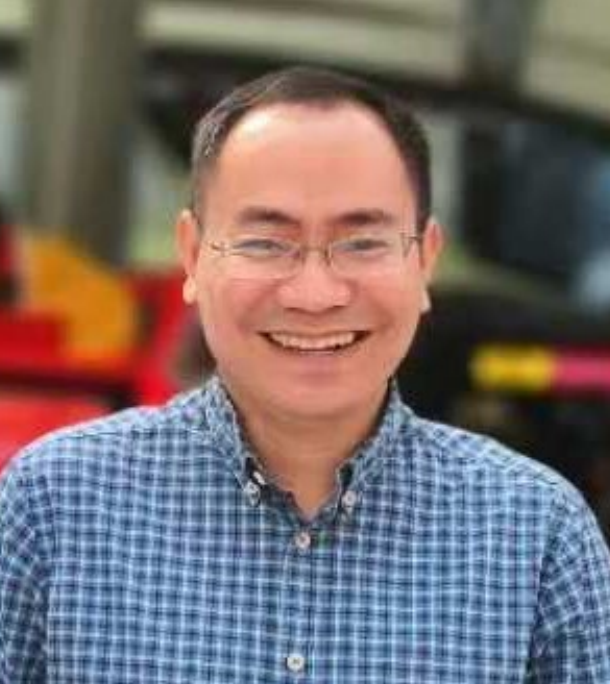}}]{Xu Chen}
received the Ph.D. degree in information engineering from the Chinese University of Hong Kong in 2012. He is a Full Professor with Sun Yat-sen University, Guangzhou, China, Director of Institute of Advanced Networking and Computing Systems, and the Vice Director of the National and Local Joint Engineering Laboratory of Digital Home. He was a Post-Doctoral Research Associate with Arizona State University, Tempe, USA, from 2012 to 2014, and a Humboldt Scholar Fellow with the Institute of Computer Science, University of Goettingen, Germany, from 2014 to 2016. He was a recipient of the Prestigious Humboldt Research Fellowship awarded by Alexander von Humboldt Foundation of Germany, the 2014 Hong Kong Young Scientist Runner-Up Award, the 2020 IEEE Computer Society Best Paper Awards Runner-Up, the 2017 IEEE Communication Society Asia–Pacific Outstanding Young Researcher Award, the 2017 IEEE ComSoc Young Professional Best Paper Award, the Honorable Mention Award of 2010 IEEE international conference on Intelligence and Security Informatics, the Best Paper Runner-Up Award of 2014 IEEE International Conference on Computer Communications (INFOCOM), and the Best Paper Award of 2017 IEEE International Conference on Communications. He is currently an Area Editor of IEEE Open Journal of the Communications Society, an Associate Editor of the IEEE Transactions Wireless Communications, IEEE Internet of Things Journal, IEEE Transactions on Vehicular Technology, and IEEE Journal on Selected Areas in Communications (JSAC) Series on Network Softwarization and Enablers.
\end{IEEEbiography}


\vskip -2\baselineskip plus -1fil
\begin{IEEEbiography}[{\includegraphics[width=1in,height=1.25in,clip,keepaspectratio]{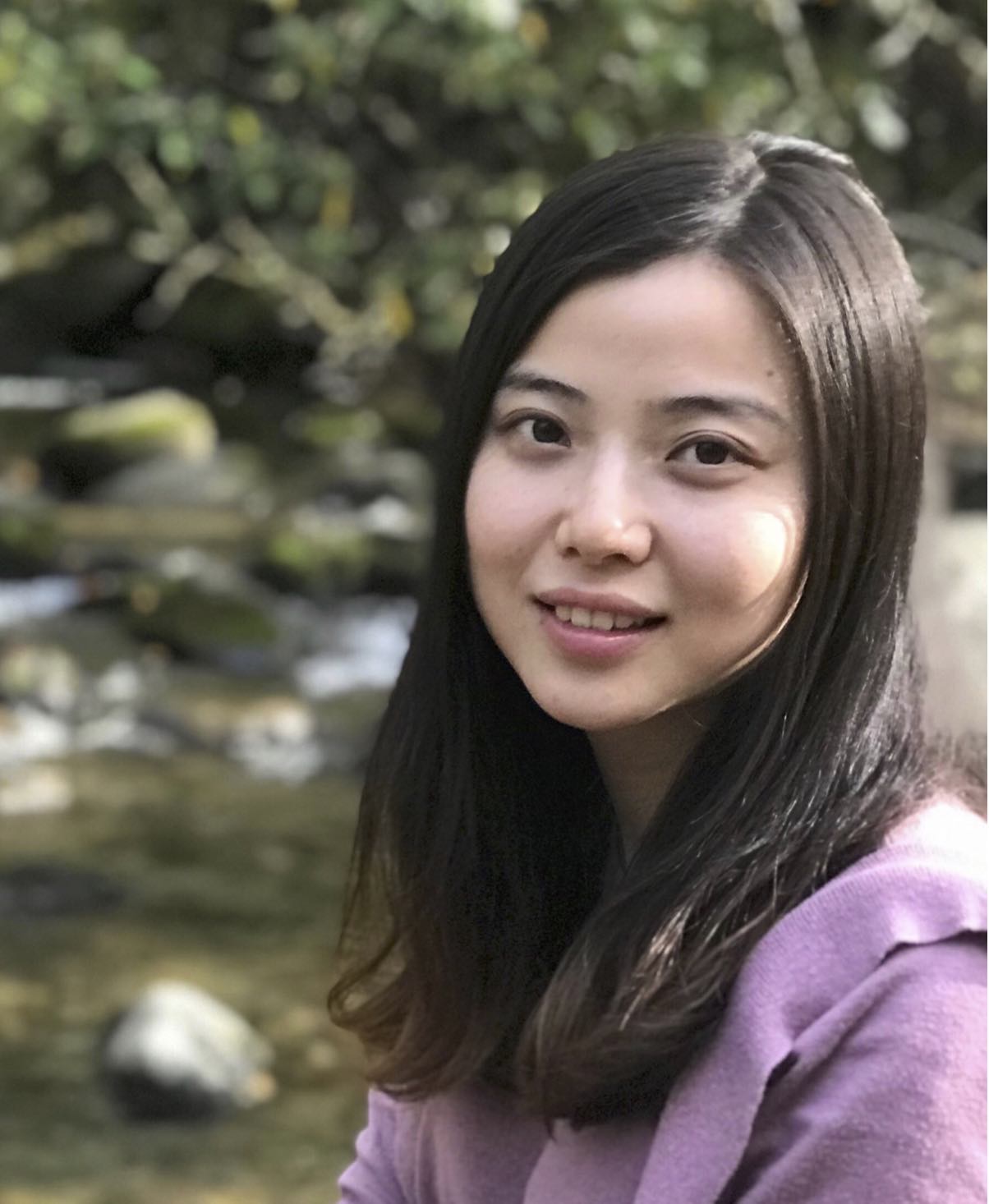}}]{Xiaoxi Zhang} received the B.E. degree in electronics and information engineering from the Huazhong University of Science and Technology in 2013 and the Ph.D. degree in computer science from The University of Hong Kong in 2017. She was a Post-Doctoral Researcher with the Department of Electrical and Computer Engineering, Carnegie Mellon University, during 2017-2020. She is currently an Associate Professor with the School of Computer Science and Engineering, Sun Yat-sen University. She was a recipient of the Young Outstanding Award of the Guangdong Province, Best Student Paper Award of IEEE MSN 2024, Best Paper Award of IEEE BigCom 2024, and Best Paper Nominee of IEEE/ACM IWQoS 2023. She is currently an Area Editor of Elsevier Computer Networks, an Associate editor of IEEE Networking Letters, and a Guest editor of MDPI Symmetry in Optimization Theory and Its Applications. She is broadly interested in optimization, algorithm design, and system implementation for networked systems, including cloud and edge computing networks, distributed machine learning systems, and vehicular networks.
\end{IEEEbiography}
\vskip -2\baselineskip plus -1fil

\end{document}